%% file: root.tex
\documentclass{article} % For LaTeX2e
\usepackage{iclr2026_conference,times}

\input{math_commands.tex}

\usepackage{hyperref}
\usepackage{url}

\definecolor{iccvblue}{rgb}{0.21,0.49,0.74}
\definecolor{darkyellow}{rgb}{0.85, 0.65, 0.0}
\definecolor{SeaGreen}{rgb}{0.180,0.545,0.341}
\usepackage{graphicx}
\usepackage{booktabs}
\usepackage{xcolor,colortbl}
\usepackage{amssymb}
\usepackage{multirow}
\usepackage{pifont}
\usepackage{wrapfig}

\usepackage{dashbox}

\definecolor{htmlcssgreen}{rgb}{0.0, 0.5, 0.0}
\newcommand{\increase}[1]{\footnotesize \textcolor{htmlcssgreen}{#1}\normalsize}

\definecolor{britishracinggreen}{rgb}{0.0, 0.26, 0.15}
\definecolor{cadmiumgreen}{rgb}{0.0, 0.42, 0.24}
\definecolor{ourpurple}{rgb}{0.40, 0.39, 1.0}
\definecolor{ForestGreen}{rgb}{0.13, 0.55, 0.13}
\definecolor{lightyellow}{HTML}{FFE6CC}
\definecolor{darkorange}{HTML}{FFB366}

\newcommand{\trisign}{\ding{115}}

\newcommand{\limited}{\textcolor{yellow}{\raisebox{-.1em}{\large 
 \trisign}}\kern-0.64em{\scriptsize \textbf{!}} \normalsize}

\newcommand{\bestresult}[1]{\boldmath{\textcolor{red}{#1}}}
\newcommand{\secondbest}[1]{\textcolor{blue}{\underline{#1}}}

\newcommand{\hlgray}[1]{\begingroup\setlength{\fboxsep}{0.8pt}\colorbox{gray!20}{#1}\endgroup}
\newcommand{\hlorange}[1]{\begingroup\setlength{\fboxsep}{0.8pt}\colorbox{orange!20}{#1}\endgroup}
\newcommand{\hlgreen}[1]{\begingroup\setlength{\fboxsep}{0.8pt}\colorbox{cadmiumgreen!20}{#1}\endgroup}

\usepackage{algorithm}
\usepackage{algpseudocode}
\usepackage{amsmath}
\usepackage{caption}
\usepackage{subcaption}
\usepackage{enumitem}

\newcommand{\algorule}{\par\noindent\rule{\linewidth}{0.4pt}\par}

\title{Cross-View Open-Vocabulary Object Detection in Aerial Imagery}

\author{
Jyoti Kini \quad
Rohit Gupta \quad
Mubarak Shah \\
Center for Research in Computer Vision, University of Central Florida, Orlando, Florida \\
{\tt\{jyoti.kini,rohit.gupta\}@ucf.edu, shah@crcv.ucf.edu}
}

\iclrfinalcopy

\begin{document}

\maketitle
\thispagestyle{fancy}
\fancyhead[L]{}

\vspace{-5mm}
\begin{abstract}
Traditional object detection models are typically trained on a fixed set of classes, limiting their flexibility and making it costly to incorporate new categories. Open-vocabulary object detection addresses this limitation by enabling models to identify unseen classes without explicit training. Leveraging pretrained models contrastively trained on abundantly available ground-view image-text classification pairs provides a strong foundation for open-vocabulary object detection in aerial imagery. Domain shifts, viewpoint variations, and extreme scale differences make direct knowledge transfer across domains ineffective, requiring specialized adaptation strategies. In this paper, we propose a novel framework for adapting open-vocabulary representations from ground-view images to solve object detection in aerial imagery through structured domain alignment. The method introduces contrastive image-to-image alignment to enhance the similarity between aerial and ground-view embeddings and employs multi-instance vocabulary associations to align aerial images with text embeddings. Extensive experiments on the xView, DOTAv2, VisDrone, DIOR, and HRRSD datasets are used to validate our approach. Our open-vocabulary model achieves improvements of +6.32 mAP on DOTAv2, +4.16 mAP on VisDrone (Images), and +3.46 mAP on HRRSD in the zero-shot setting when compared to finetuned closed-vocabulary dataset-specific model performance, thus paving the way for more flexible and scalable object detection systems in aerial applications.
\end{abstract}

\vspace{-5mm}
\section{Introduction}
\label{sec:intro}
\vspace{-3mm}

% \begin{figure}[t]
% \begin{center}
% \includegraphics[width=0.5\linewidth]{images/Teaser.png}
% \end{center}
%    \caption{\textbf{Motivation} 
%    Conventional ground-view contrastive pretraining (top-left) fails to generalize to aerial views due to modality gaps, and traditional aerial finetuning (top-right) results in misaligned feature spaces. Our method (bottom-left) enforces cross-view contrastive alignment and aerial-text associations, ensuring better semantic consistency and generalization for open-vocabulary aerial object detection (bottom-right). \underline{\textit{Note}}: Different object classes are color-coded: \textcolor{red}{Airplane}, \textcolor{SeaGreen}{Car}, \textcolor{darkyellow}{Truck}. Different domain embeddings are shape-coded: $\blacktriangle$ Ground-view, $\blacksquare$ Text, $\bigstar$ Aerial-view.}
% \label{fig:Speed}
% \end{figure}

Over the last decade, object detection has seen remarkable progress, with models attaining high accuracy on predefined categories in closed-set settings \cite{he2017mask, ren2015faster, carion2020end}. However, real-world applications demand open-vocabulary object detection, where models must recognize and localize unseen object categories. The emergence of large-scale pretrained Vision-Language Models (VLMs) \cite{radford2021learning, jia2021scaling} has significantly advanced open-vocabulary object detection, enabling models to detect novel categories beyond their training set. Among these, contrastive vision-language models \cite{ranasinghe2023perceptual, khan2023contrastive, ak2024aligning, cui2022contrastive} such as OWLv2 \cite{minderer2023scaling} leverage large-scale image-text pairings to learn generalizable representations, facilitating zero-shot object detection and retrieval. While these models achieve strong performance in ground-view imagery, extending them to aerial-view object detection remains a fundamental challenge due to domain shifts and viewpoint variations. Aerial imagery introduces unique complexities: extreme scale variations, occlusions, and drastic perspective distortions, making direct transfer of ground-view models ineffective.

Researchers have developed ground-to-aerial domain adaptation techniques to leverage the abundance of labeled ground-level image data for aerial object detection. Existing approaches address the domain gap through generative modeling \cite{zeng2024unsupervised, ma2024decomposition, soto2020domain, ye2024skydiffusion, lu2020geometry}, adversarial training \cite{chen2021targeted, wozniak2024uada3d}, self-supervised learning \cite{scheibenreif2024parameter, wang2024siamseg}, viewpoint-dependent feature matching \cite{mule2025enhancing, hou2023feature, shan2014accurate, shugaev2024arcgeo, regmi2019bridging}, and knowledge distillation \cite{yao2024domain, wang2024multi}. 

Generative models, such as GANs, synthesize aerial-like images from ground-view data, expanding training datasets and enhancing model generalization. Adversarial domain adaptation reduces domain shift by training a discriminator to distinguish between ground and aerial feature distributions while guiding the model to learn domain-invariant representations. Self-supervised learning leverages the structure of unlabeled aerial images through pretext tasks like rotation prediction, enabling the model to learn useful representations without explicit supervision. Viewpoint-dependent feature 
\begin{wrapfigure}{r}{0.47\linewidth}
% \begin{figure}
\vspace{-1em}
\begin{center}
\includegraphics[width=\linewidth]{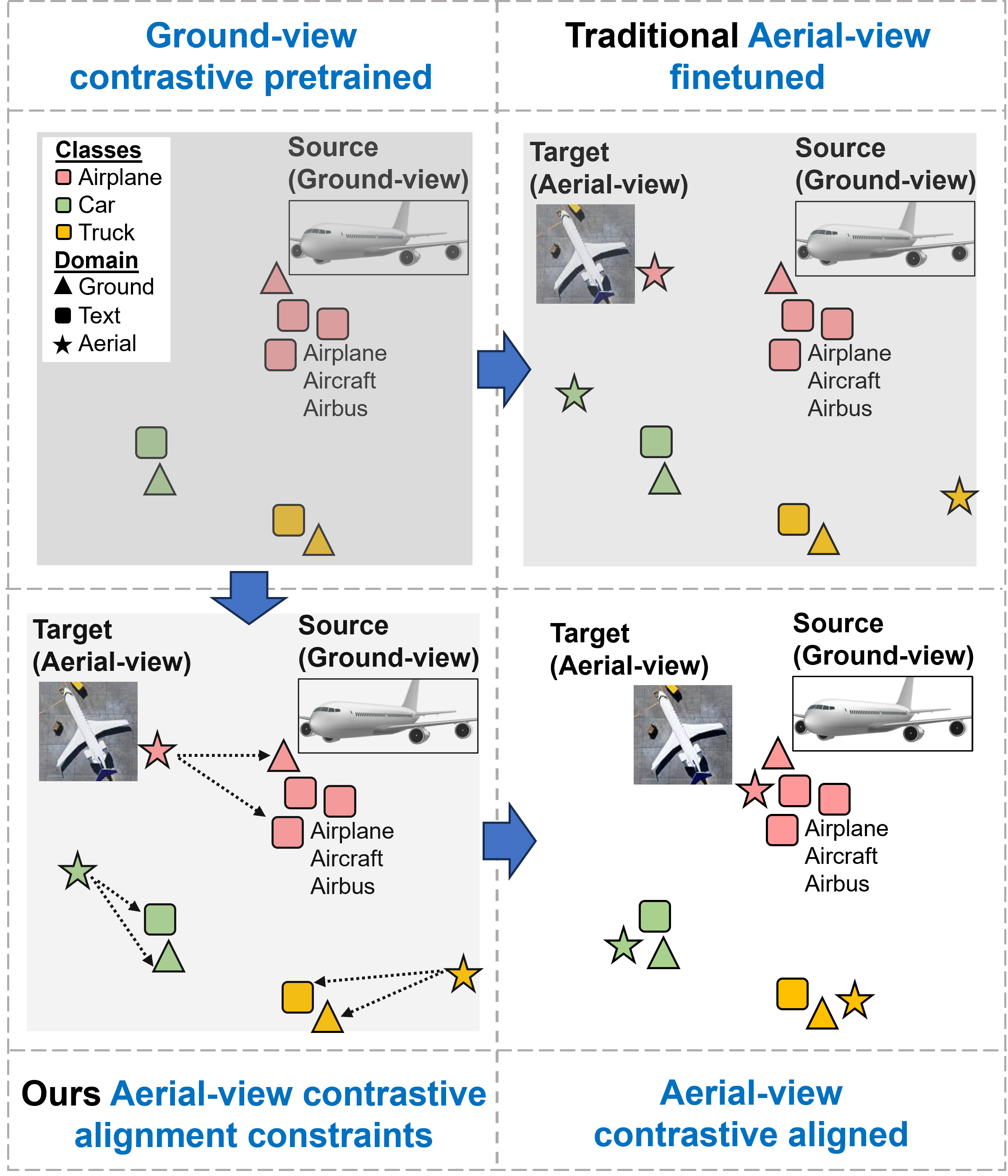}
\end{center}
\vspace{-1em}
\caption{\textbf{Motivation:}
Ground-view contrastive pretrained open vocabulary detectors (top-left) fail to generalize to aerial views, and traditional finetuning (top-right) results in misaligned feature spaces. Our method (bottom-left) enforces cross-view contrastive alignment and aerial view-text associations, ensuring better semantic consistency and generalization for open-vocabulary aerial object detection (bottom-right). \underline{\textit{Note}}: Different object classes are color-coded: \textcolor{red}{Airplane}, \textcolor{SeaGreen}{Car}, \textcolor{darkyellow}{Truck}. Different domains are shape-coded: $\blacktriangle$ Ground-view, $\blacksquare$ Text, $\bigstar$ Aerial-view.}
\vspace{-1.5em}
\label{fig:teaser}
% \end{figure}
\end{wrapfigure}
matching establishes correspondences between ground and aerial imagery by leveraging geometric relationships, ensuring better feature alignment. Lastly, knowledge distillation transfers information from a ground-view teacher model to a student model optimized for aerial detection, facilitating effective cross-domain knowledge transfer. 

% % \begin{wrapfigure}{r}{0.47\linewidth}
% \begin{figure}
% % \vspace{-2.5em}
% \begin{center}
% \includegraphics[width=0.36\linewidth]{images/Teaser.png}
% \includegraphics[width=0.55\linewidth]{images/motivation_by_model_ochre_skyblue_centerLegend_arrow_drop_noDown.pdf}
% \end{center}
% \vspace{-1em}
% \caption{\textbf{Motivation}
% Conventional ground-view contrastive pretraining (top-left) fails to generalize to aerial views due to modality gaps, and traditional aerial finetuning (top-right) results in misaligned feature spaces. Our method (bottom-left) enforces cross-view contrastive alignment and aerial-text associations, ensuring better semantic consistency and generalization for open-vocabulary aerial object detection (bottom-right). \underline{\textit{Note}}: Different object classes are color-coded: \textcolor{red}{Airplane}, \textcolor{SeaGreen}{Car}, \textcolor{darkyellow}{Truck}. Different domains are shape-coded: $\blacktriangle$ Ground-view, $\blacksquare$ Text, $\bigstar$ Aerial-view.}
% \vspace{-1em}
% \label{fig:Speed}
% \end{figure}
% % \end{wrapfigure}

Beyond domain adaptation, research in open-vocabulary aerial detection aims to recognize objects outside the set of categories encountered during training. CastDet \cite{li2024toward} utilizes a self-learning framework based on a student-teacher paradigm, integrating RemoteCLIP \cite{liu2024remoteclip} to generate class-agnostic region proposals and pseudo-labels, thereby enabling open-vocabulary detection in aerial imagery. Another approach, SS-OWFormer \cite{mullappilly2024semi}, designed for satellite imagery, employs a feature-alignment mechanism alongside pseudo-labeling to enhance the detection of previously unseen objects. However, these methods heavily depend on large-scale aerial datasets for finetuning, pseudo-labeling, or feature alignment, making them reliant on labeled aerial supervision. In practice, such large-scale labeled datasets are often scarce or unavailable in the aerial domain, which further limits the applicability of these approaches.

\begin{wrapfigure}{r}{0.43\textwidth}
\vspace{-2em}
  \centering
  \vspace{-0.5\baselineskip} % tighten top spacing (optional)
  \includegraphics[width=\linewidth]{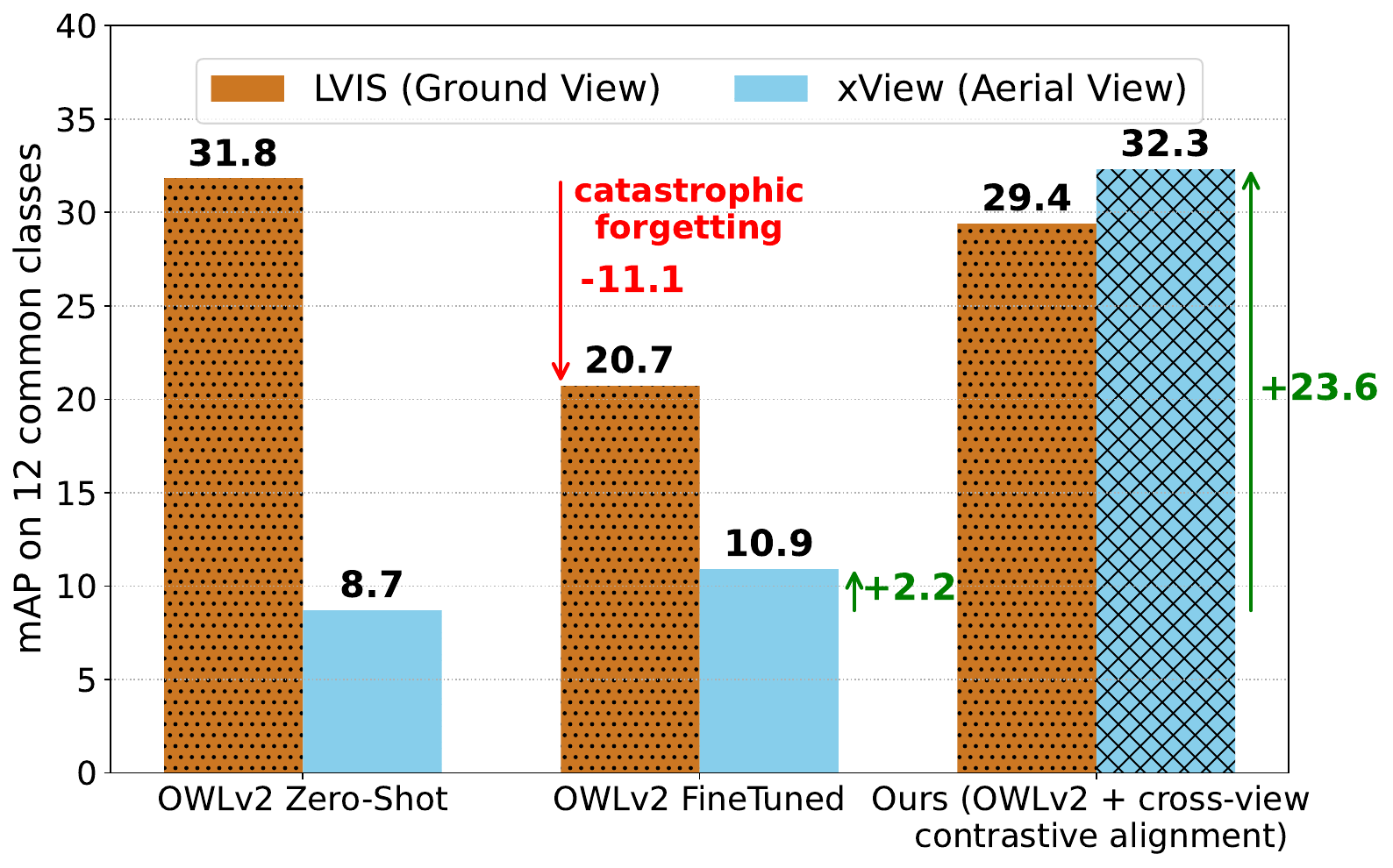}
  \vspace{-2em}
  \caption{Our approach improves aerial-view detection while preserving ground-view performance and avoids catastrophic forgetting typical of naive finetuning.}
  \vspace{-1.5em}
  \label{fig:motivation_quant}
\end{wrapfigure}

% \begin{figure}[h]
%     \centering
%     \includegraphics[width=0.5\linewidth]{images/motivation_v2_fonts24.pdf}
%     \caption{Caption}
%     \label{fig:motivation_quant}
% \end{figure}

In contrast, we propose a novel approach leveraging the abundance of ground-view image-text data available and align it with aerial representations using a contrastive learning framework, as seen in Fig.~\ref{fig:teaser}. Unlike approaches that require dataset-specific finetuning, which can lead to catastrophic forgetting of ground-view knowledge (refer Fig.~\ref{fig:motivation_quant}), our contrastive learning formulation preserves pretrained vision-language model knowledge while ensuring robust cross-view feature alignment. Our vocabulary expansion enhances zero-shot detection of novel objects, making it more scalable and adaptable for open-vocabulary aerial object detection compared to existing approaches.

% \noindent
% The key contributions of this work are as follows:
% \begin{itemize}[nosep,leftmargin=*]
%     \item \textbf{Contrastive Image-to-Image Alignment:} We introduce a novel contrastive alignment strategy that enhances semantic consistency between aerial-view and ground-view embeddings, facilitating the transfer of meaningful features from ground-view pretrained vision-language models (VLMs) to open-vocabulary object detection in aerial imagery. \vspace{2mm}
 
%     \item \textbf{Multi-Instance Vocabulary Association:} We propose a multi-instance vocabulary association technique that aligns aerial images with text embeddings, effectively leveraging textual descriptions to improve recognition of unseen categories in aerial view images. \vspace{2mm}
    
%     \item \textbf{Extensive Benchmark Evaluation:} We validate  experiments on the xView \cite{lam2018xview}, DOTAv2 \cite{xia2018dota}, VisDrone \cite{cao2021visdrone}, DIOR \cite{li2020object}, and HRRSD \cite{zhang2019hierarchical} datasets, achieving significant improvements in open-vocabulary detection performance, highlighting the scalability and effectiveness of our method.
% \end{itemize}

\noindent
The key contributions of this work are as follows: \textbf{a. Contrastive Image-to-Image Alignment:} We introduce a novel contrastive alignment strategy that enhances semantic consistency between aerial-view and ground-view embeddings, facilitating the transfer of meaningful features from ground-view pretrained vision-language models (VLMs) to open-vocabulary object detection in aerial imagery. \textbf{b. Multi-Instance Vocabulary Association:} We propose a multi-instance vocabulary association technique that aligns aerial images with text embeddings, effectively leveraging textual descriptions to improve recognition of unseen categories in aerial view images. \textbf{c. Extensive Benchmark Evaluation:} We validate  experiments on the xView \cite{lam2018xview}, DOTAv2 \cite{xia2018dota}, VisDrone \cite{cao2021visdrone}, DIOR \cite{li2020object}, and HRRSD \cite{zhang2019hierarchical} datasets, achieving significant improvements in open-vocabulary detection performance, highlighting the scalability and effectiveness of our method.

% \noindent
% The key contributions of this work are as follows: \textbf{a. Contrastive Image-to-Image Alignment:} We introduce a novel contrastive alignment strategy that enhances semantic consistency between aerial-view and ground-view embeddings, facilitating the transfer of meaningful features from ground-view pretrained vision-language models (VLMs) to open-vocabulary object detection in aerial imagery. \textbf{b. Multi-Instance Vocabulary Association:} To enhance the adaptability of VLMs to aerial imagery, we propose a multi-instance vocabulary association technique that aligns aerial images with text embeddings, effectively leveraging textual descriptions to improve recognition of unseen categories. \textbf{c. Extensive Benchmark Evaluation:} We validate our approach through rigorous experiments on the xView \cite{lam2018xview}, DOTAv2 \cite{xia2018dota}, VisDrone \cite{cao2021visdrone}, DIOR \cite{li2020object}, and HRRSD \cite{zhang2019hierarchical} datasets, demonstrating significant improvements in open-vocabulary object detection performance, highlighting the scalability and effectiveness of our method.

\vspace{-3mm}
\section{Related Work}
\vspace{-3mm}
\noindent
\textbf{Aerial Object Detection} Deep learning has significantly advanced object detection in natural/ground-view images, leading to the adaptation of frameworks like Faster R-CNN \cite{girshick2015fast}, RetinaNet \cite{lin2017focal}, YOLO \cite{redmon2016you}, and DETR \cite{zhu2020deformable} for aerial imagery. However, aerial object detection faces challenges such as extreme scale variations, arbitrary orientations, and high object density. To address these challenges, methods like ROI-Transformer, R3Det \cite{yang2021r3det}, and RSDet++ \cite{qian2022rsdet} improve localization accuracy by using oriented bounding boxes, while techniques such as SCRDet \cite{yang2019scrdet} and ClusDet \cite{yang2019clustered} enhance the detection of small and densely packed objects. Frameworks like Focus-and-Detect \cite{koyun2022focus} and PARE-YOLO \cite{zhang2025fusion} further refine small object detection through clustering and multi-scale feature fusion. NavBLIP \cite{li2025navblip} integrates visual and contextual information for UAV-based detection, enhancing performance in dynamic environments. Despite these advancements, most methods operate under a closed-set assumption, meaning they are trained and evaluated on a fixed set of object categories. Expanding such models to recognize novel categories requires collecting large-scale annotated datasets, a costly and time-intensive process. 

\noindent
\textbf{Open-Vocabulary Object Detection} (OVD) Early work such as OVR-CNN \cite{zareian2021open} used bounding box annotations for a limited set of objects while leveraging image-caption pairs to extend the detection vocabulary. With the advent of pretrained Vision-Language Models (VLMs) like CLIP \cite{radford2021learning} and ALIGN \cite{jia2021scaling}, recent OVD approaches have transferred text-image knowledge into detection models using prompt learning and region-level finetuning. Methods like ViLD \cite{gu2021open} employ vision-language distillation to align textual and visual features, RegionCLIP \cite{zhong2022regionclip} aligns region-level representations with textual concepts, and Detic \cite{zhou2022detecting} enhances detection capabilities by incorporating large-scale image classification datasets. Further refinements, including PromptDet \cite{feng2022promptdet} and DetPro \cite{du2022learning}, optimize prompt embeddings for better visual-textual alignment. Additionally, models like YOLO-World \cite{cheng2024yolo} and F-VLM \cite{kuo2022f} address open-vocabulary detection challenges. YOLO-World integrates real-time vision-language modeling with Ultralytics YOLOv8, enabling efficient object detection based on descriptive text. F-VLM simplifies the training process by using frozen vision and language models, achieving state-of-the-art results with reduced computational overhead. Despite these advancements, applying OVD to aerial imagery remains challenging due to smaller dataset sizes and fundamental differences from ground-view images.

\begin{figure*}
  \centering
  \vspace{2mm}
  \includegraphics[width=1\textwidth]{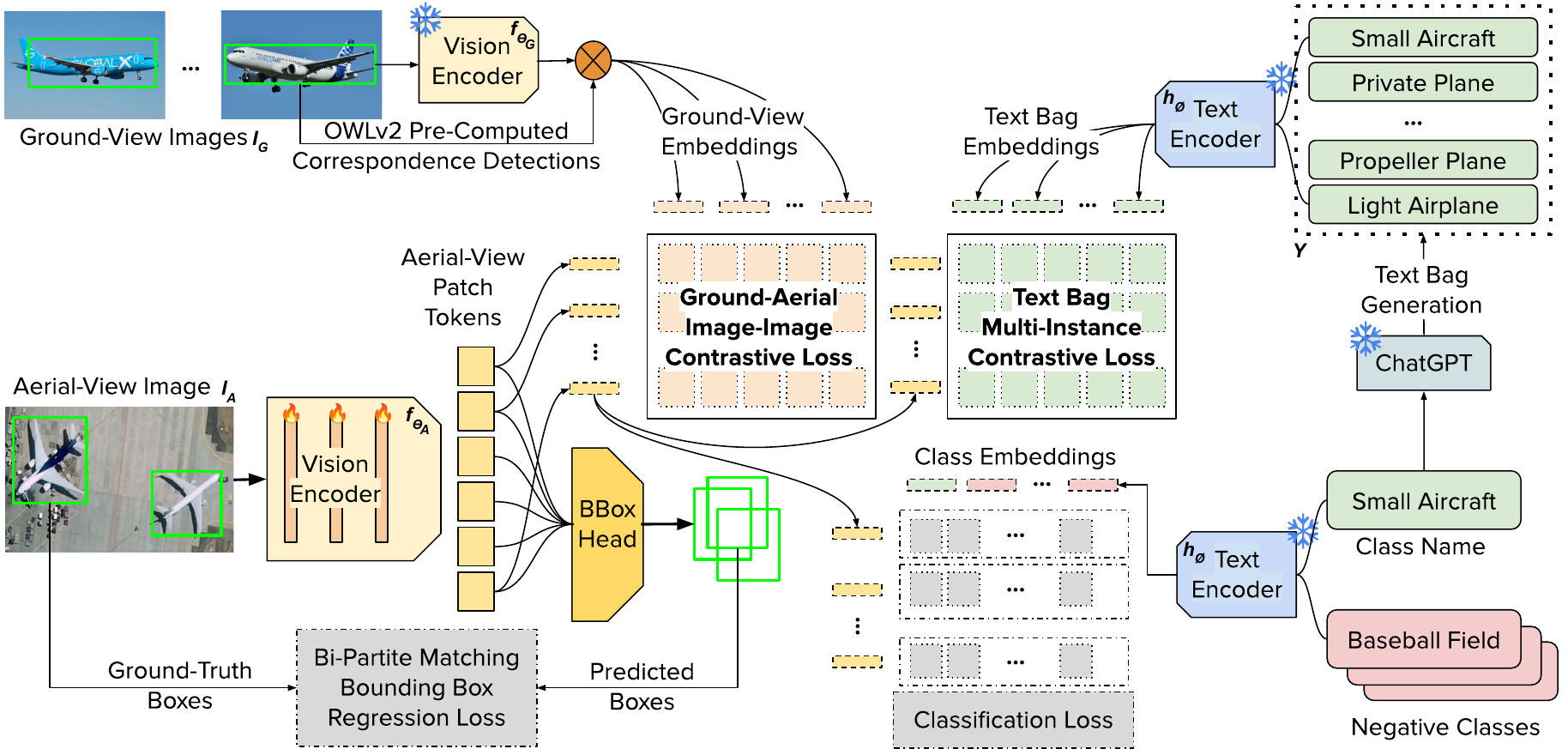}
  \vspace{-4mm}
\caption{\textbf{Overview} 
Traditional open-vocabulary object detector finetuning includes two losses: a bi-partite matching based {\protect\hlgray{bounding box regression loss}} and a {\protect\hlgray{classification loss}} computed in the shared image–text embedding space. We further introduce two new components: a {\protect\hlorange{ground–aerial contrastive loss}} that aligns aerial and ground image embeddings, and a {\protect\hlgreen{text-bag multi-instance contrastive loss}} that aligns aerial features with text bags. To realize these objectives, we generate an aerial–ground correspondence dataset providing cross-view positives/negatives, and also expand the class vocabulary by using ChatGPT to synthesize variants of class names (“text bags”) consumed by the text encoder. Joint optimization over these losses places aerial $\leftrightarrow$ ground $\leftrightarrow$ text representations in a cohesive shared embedding space, yielding stronger zero-shot generalization in aerial imagery.
}
  \label{fig:architecture}
  \vspace{-3mm}
\end{figure*}

\vspace{-3mm}
\section{Approach}
\label{sec:approach}
\vspace{-3mm}
Recent advancements in vision-language models (VLMs) have significantly improved open-vocabulary object detection by aligning images and text in a shared embedding space. Our framework builds upon this idea to tackle cross-view open-vocabulary object detection, specifically aligning aerial-view images with ground-view and textual representations. Our model consists of a vision encoder \( f_\theta \) and a text encoder \( h_\phi \), which respectively project aerial images and textual queries into a common feature space. Given an aerial image \( I_A \), the vision encoder extracts its feature representation \( f_\theta(I_A) \in \mathbb{R}^d \), where \( d \) is the size of the embedding space.  Similarly, a textual query \( Y \) is tokenized and processed by the text encoder \( h_\phi \), producing a textual feature embedding \( h_\phi(Y) \in \mathbb{R}^d \).  

To ensure effective alignment between aerial, ground, and textual representations, our model incorporates two key learning mechanisms: Cross-View Representation Alignment (Sec.~\ref{sec:cross_view_alignment}) and Aerial-Text Multi-Instance Association (Sec.~\ref{sec:aerial_text_association}). The success of these alignment strategies depends on the quality of the generated contrastive alignment data. To construct this, we introduce a data generation pipeline comprising Aerial-Ground Object Detection Correspondence (Sec.~\ref{sec:aerial_ground_corr}) and Aerial-Text Vocabulary Expansion (Sec.~\ref{sec:aerial_text_vocab}).

\vspace{-3mm}
\subsection{Cross-View Representation Alignment}
\label{sec:cross_view_alignment}
\vspace{-3mm}
Standard contrastive vision-language models struggle to generalize from ground-view pretraining to aerial views due to the modality gap introduced by viewpoint shifts and domain discrepancies. Instead of direct finetuning on aerial imagery, we propose a cross-view contrastive alignment strategy that explicitly bridges the feature spaces of aerial and ground representations. 

Given a batch of aerial images \( I_A \) and corresponding ground-view references \( I_G \), we apply contrastive learning to enforce similarity between aerial-ground pairs, while simultaneously pushing apart non-matching samples. Since only the aerial-view encoder is finetuned during training, we use a single notation \( f_{\theta} \) to refer to the model being optimized, simplifying the formulation:
\vspace{-2mm}
\begin{equation}
\label{eq:cross_view}
\mathcal{L}_{\textit{\textcolor{darkyellow}{Image\textsubscript{A}}--\textcolor{orange}{Image\textsubscript{G}}}} = -\frac{1}{N} \sum_{i=1}^{N} \log \frac{\exp(f_\theta(I_A^i)^\top, f_\theta(I_G^i)) / \rho}
{\sum_{j=1}^{N} \exp(f_\theta(I_A^i)^\top, f_\theta(I_G^j)) / \rho},
\end{equation}
\vspace{-0.5mm}
where \( N \) represents the batch size, \( i \) indexes a positive (correctly matched) aerial-ground pair, and \( j \) indexes all samples in the batch. The numerator maximizes the similarity between the correct aerial-ground pair \( (I_A^i, I_G^i) \), while the denominator contrasts it against all ground-view images \( I_G^j \) in the batch, encouraging better separation between matching and non-matching samples. Here, \( \rho \) is a temperature parameter controlling the sharpness of the similarity distribution.

By omitting a subscript for \( f_{\theta} \), we emphasize that only one encoder is being finetuned (i.e., the aerial-view encoder), while the ground-view embeddings remain fixed, ensuring efficient adaptation of aerial representations to the pretrained feature space.

\vspace{-3mm}
\subsection{Aerial-Text Multi-Instance Association}
\label{sec:aerial_text_association}
\vspace{-3mm}
To further improve generalization, we introduce multi-instance contrastive learning between aerial images and their corresponding textual descriptions. Rather than enforcing a strict one-to-one correspondence, we adopt a multi-instance text association approach, where each aerial image \( I_A^i \) is linked to a set of semantically related textual queries via a text-bag representation. This formulation enables better alignment between aerial imagery and natural language descriptions by accounting for ambiguities and variations in object naming conventions.

Our objective function extends MIL-NCE \cite{miech2020end}, optimizing the alignment between aerial images and a diverse set of textual candidates:
\vspace{-2mm}
\begin{equation}
\label{eq:multi_instance}
\mathcal{L}_{\textit{\textcolor{darkyellow}{Image\textsubscript{A}}--\textcolor{cadmiumgreen}{Text}}} = -\frac{1}{N} \sum_{i=1}^{N} \log \frac{\sum_{k} \sum_{n} \exp(h_{\phi}(Y_{k,n})^\top f_\theta(I_A^i)/\sigma)}
{\sum_{j} \sum_{n} \exp(h_{\phi}(Y_{j,n})^\top f_\theta(I_A^i)/\sigma)},
\end{equation}
\vspace{-0.5mm}
where \( N \) represents the batch size, index \( i \) refers to an aerial image \( I_A^i \), while \( j \) iterates over all possible textual descriptions in the batch. The index \( k \) corresponds to a positive text-bag group, which contains multiple semantically related textual descriptions for \( I_A^i \), and \( n \) indexes the individual text descriptions within the text-bag. The numerator sums over multiple positive textual descriptions \( Y_{k,n} \) associated with \( I_A^i \), while the denominator contrasts it against all text samples \( Y_{j,n} \) in the batch, ensuring that the model learns to distinguish between relevant and irrelevant textual associations. This multi-instance contrastive loss promotes robust open-vocabulary object detection.

\vspace{-5mm}
\subsection{Aerial-Ground Object Detection Correspondence}
\label{sec:aerial_ground_corr}
\vspace{-3mm}
% \begin{figure}
%   \centering
%   \vspace{2mm}
%   \includegraphics[width=0.5\linewidth]{images/DataGeneration.png}
%   \vspace{-3mm}
% \caption{\textbf{Aerial-Ground Object Detection Correspondence} 
% Step 1 (top-left) shows a case where the category `Car' is annotated in both aerial and ground-view datasets, allowing direct positive pair generation using ground-truth annotations. Step 2 (top-right) illustrates a case where the category `Building' lacks ground-view annotations. To address this, OWLv2 is used to generate detections in the ground-view image (bottom row), followed by non-maximum suppression to establish cross-view correspondence data for our contrastive alignment training.}
%   \label{fig:data_generation_1}
%   \vspace{-3mm}
% \end{figure} 

\begin{wrapfigure}{r}{0.65\linewidth}
\centering
\vspace{-1.5em}
\includegraphics[width=\linewidth]{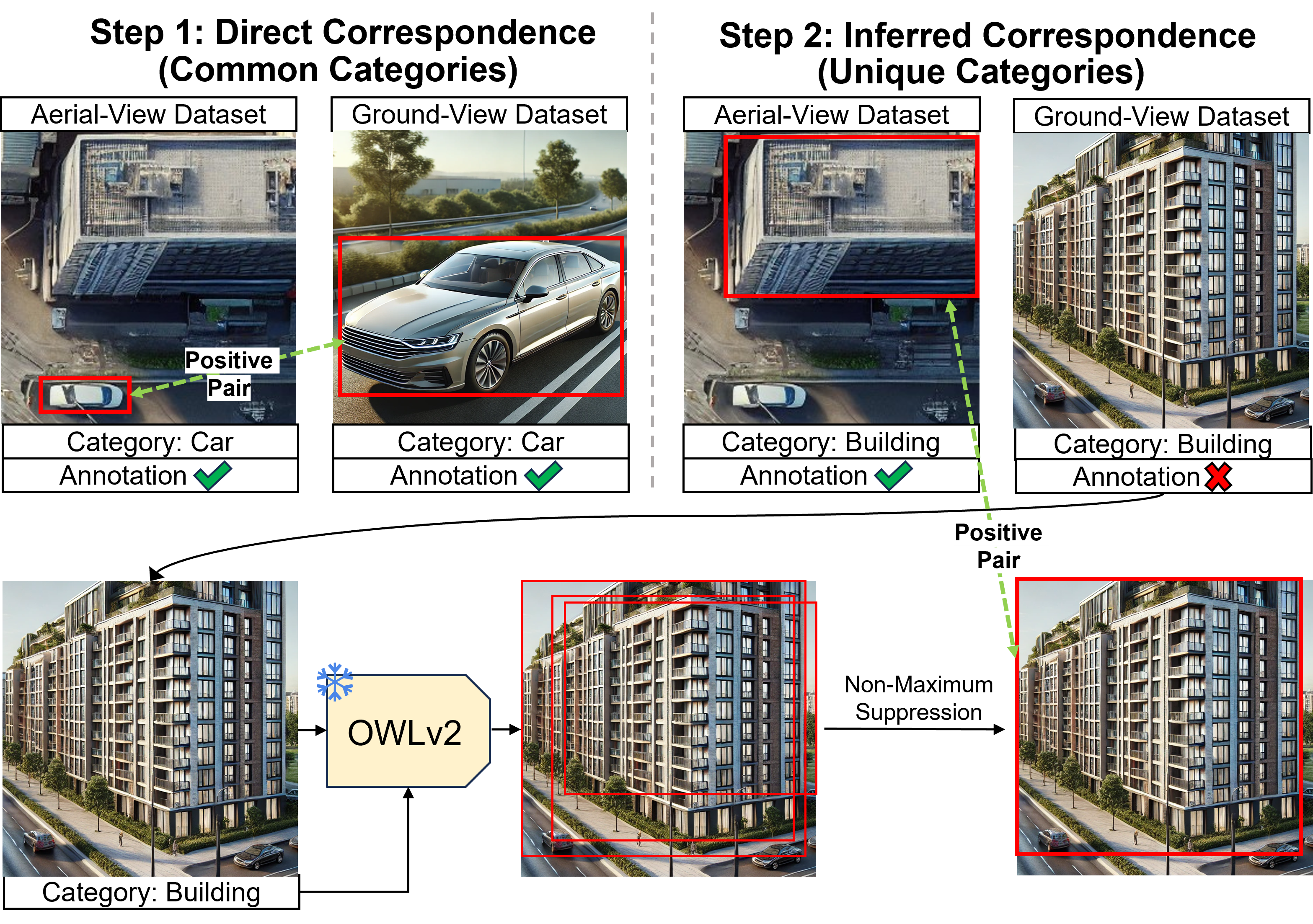}
\vspace{-2em}
\caption{\textbf{Aerial-Ground Object Detection Correspondence}
Step 1 (top-left) shows a case where the category `Car' is annotated in both aerial and ground-view datasets, allowing direct positive pair generation using ground-truth annotations. Step 2 (top-right) illustrates a case where the category `Building' lacks ground-view annotations. To address this, OWLv2 is used to generate detections in the ground-view image (bottom row), followed by non-maximum suppression to establish cross-view correspondence data for our contrastive alignment training.}
\label{fig:data_generation_1}
\vspace{-0.7em}
\end{wrapfigure}

In cross-view object detection, establishing category-based correspondence between aerial and ground-view detections is crucial for model generalization. Our data generation pipeline constructs aerial-ground correspondences by leveraging existing datasets (namely xView \cite{lam2018xview} for aerial and LVIS \cite{gupta2019lvis} / CC12M \cite{changpinyo2021conceptual} for ground) and enhancing them with inferred-labeling techniques. The goal is to align aerial detections with semantically corresponding ground-view samples while ensuring sufficient diversity. To facilitate \vspace{-2mm}

cross-view representation learning for open-vocabulary aerial object detection, we construct a dataset \( D_{aligned} \) by establishing correspondences between aerial and ground-view images. This process consists of two primary steps: \textbf{Direct Correspondence}, as shown in Fig.~\ref{fig:data_generation_1}, where detection matches between aerial and ground images are identified based on category, and \textbf{Inferred Correspondence}, where pseudo-matches are generated for categories that lack ground-view annotations. The final step involves \textbf{Augmentation and Normalization} to enhance the generated data’s diversity and consistency.

\captionsetup[algorithm]{skip=0pt} % Adjusts space between algorithm and caption

\begin{wrapfigure}[20]{r}{0.63\linewidth} 
\vspace{-0.5em}
% \hspace{-1em}
  \vspace{-\intextsep}
  \vspace{0.5ex}\noindent\rule{\linewidth}{1.2pt}
  \begin{minipage}{\linewidth}
    \captionsetup{type=algorithm}
    \caption{Aerial-Ground Object Detection Correspondence}
    \label{alg:correspondence_generation}

    \algorule
    \begin{algorithmic}[1]
        \State \textbf{Initialize:} $D_{aligned} \gets \emptyset$ \hfill \textcolor{gray}{\# Empty dataset}
        
        \vspace{0.2em}
        \Statex \hspace{-2mm} $\triangleright$ \textbf{Step 1: Direct Correspondence (Common Categories)}
        \For{$c \in C_{common}$}
            \State Extract $\{B_G, I_G\}$  \hfill \textcolor{gray}{\# Ground Boxes, Images}
            \State $D_{aligned} \gets D_{aligned} \cup \{(B_A, B_G, y_c)\}$
        \EndFor
        
        \vspace{0.2em}
        \Statex \hspace{-2mm} $\triangleright$ \textbf{Step 2: Inferred Correspondence (Unique Categories)}
        \For{$c \in C_{unique}$}
            \State $\hat{B}_G \gets \text{OWLv2}(I_G)$ \hfill \textcolor{gray}{\# Generate ground boxes}
            \State $\hat{B}_G \gets \text{NMS}(\hat{B}_G, \tau)$ \hfill \textcolor{gray}{\# Remove redundant}
            \State $D_{aligned} \gets D_{aligned} \cup \{(B_A, \hat{B}_G, y_c)\}$
        \EndFor
        
        \vspace{0.2em}
        \Statex \hspace{-2mm} $\triangleright$ \textbf{Step 3: Augmentation \& Normalization}
        \State $D_{aligned} \gets \text{Augment}(D_{aligned})$ \hfill \textcolor{gray}{\# Crop, Rotate}
        \State $D_{aligned} \gets \text{Norm}(D_{aligned})$ \hfill \textcolor{gray}{\# Feature Align}
        \State \textbf{Return} $D_{aligned}$ \hfill \textcolor{gray}{\# Store for training}
        \vspace{-1em}
    \end{algorithmic}
    \algorule
  \end{minipage}
  \vspace{-\intextsep}
  \vspace{-2em}
\end{wrapfigure}

\noindent\textbf{Direct Correspondence (For Common Categories)}: For object categories present in both the aerial and ground datasets, we directly extract aligned object instances. As seen in Algo.~\ref{alg:correspondence_generation}, for each category \( c \) in the set of common categories \( C_{common} \), we retrieve bounding box annotations \( B_G \) and corresponding images \( I_G \) from the ground-view dataset. These ground-view instances are then paired with their aerial counterparts, forming structured dataset entries \( (B_A, B_G, y_c) \), where \( B_A \) represents the aerial bounding box, \( B_G \) is the corresponding ground bounding box, and \( y_c \) is the class label. This process ensures precise alignment between aerial and ground perspectives, establishing a strong foundation for  shared representations. 

\noindent\textbf{Inferred Correspondence (For Unique Categories)}: For categories that exist in the aerial dataset but lack explicit ground-view annotations, we generate inferred correspondences using vision-language models. For each category \( c \in C_{unique} \), we apply OWLv2 to detect objects in the ground-view dataset, producing an initial set of bounding boxes \( \hat{B}_G \). To refine these detections, we employ Non-Maximum Suppression (NMS) with a confidence threshold \( \tau \) to eliminate redundant or low-confidence detections, ensuring only reliable bounding boxes are retained. The final pseudo-ground view detections \( \hat{B}_G \) are then associated with aerial objects, forming dataset pairs \( (B_A, \hat{B}_G, y_c) \). This inferred alignment enables training on categories that lack explicit ground-truth annotations.

\noindent\textbf{Augmentation and Normalization}: To enhance dataset robustness and improve alignment consistency, we apply a series of augmentation and normalization techniques. Geometric transformations, such as cropping and rotation, diversify the dataset and mitigate overfitting. Additionally, feature normalization is applied to harmonize aerial and ground embedding spaces, ensuring better compatibility during model training. After these refinements, the final dataset \( D_{\text{aligned}} \) is stored for training.

\vspace{-3mm}
\subsection{Aerial-Text Vocabulary Expansion}
\label{sec:aerial_text_vocab}
\vspace{-3mm}
% \begin{wrapfigure}{r}{0.52\textwidth} % r or l, adjust width as needed
%   \vspace{-\baselineskip}             % optional: pull it up to align with text
%   \captionsetup{type=algorithm}       % make the caption count as an algorithm

%   \begin{algorithmic}[1]
%       \State \textbf{Initialize:} $D_{text} \gets \emptyset$ \hfill \textcolor{gray}{\# Empty dataset}
%       \For{$c \in C_{A}$}
%           \State $\mathcal{V}_c \gets \text{ChatGPT}(c)$ \hfill \textcolor{gray}{\# Generate label variations}
%           \State $D_{text} \gets D_{text} \cup \{(c, \mathcal{V}_c)\}$
%       \EndFor
%       \State \textbf{Return} $D_{text}$ \hfill \textcolor{gray}{\# Store for training}
%   \end{algorithmic}

%   \caption{Aerial-Text Vocabulary Expansion}
%   \label{alg:multi_instance_association}
% \end{wrapfigure}

% \begin{algorithm}
% \caption{Aerial-Text Vocabulary Expansion}
% \label{alg:multi_instance_association}
% \begin{algorithmic}[1]
%     \State \textbf{Initialize:} $D_{text} \gets \emptyset$ \hfill \textcolor{gray}{\# Empty dataset}
%     \For{$c \in C_{A}$}
%         \State $\mathcal{V}_c \gets \text{ChatGPT}(c)$ \hfill \textcolor{gray}{\# Generate label variations}
%         \State $D_{text} \gets D_{text} \cup \{(c, \mathcal{V}_c)\}$
%     \EndFor
    
%     \State \textbf{Return} $D_{text}$ \hfill \textcolor{gray}{\# Store for training}
% \end{algorithmic}
% \end{algorithm}

\begin{wrapfigure}[11]{r}{0.45\linewidth}
\vspace{-1em}
  \vspace{-\intextsep}
  \vspace{0.5ex}\noindent\rule{\linewidth}{1.2pt}
  \begin{minipage}{\linewidth}
    \captionsetup{type=algorithm}
    \caption{Vocabulary Expansion}
    \label{alg:multi_instance_association}
    \algorule
    \begin{algorithmic}[1]
        \State \textbf{Initialize:} $D_{\text{text}} \gets \emptyset$ \hfill \textcolor{gray}{\# Empty dataset}
        \For{$c \in C_{A}$}
            \State $\mathcal{V}_c \gets \text{ChatGPT}(c)$ \hfill \textcolor{gray}{\# Generate label variations}
            \State $D_{\text{text}} \gets D_{\text{text}} \cup \{(c, \mathcal{V}_c)\}$
        \EndFor
        \State \textbf{Return} $D_{\text{text}}$ \hfill \textcolor{gray}{\# Store for training}
    \end{algorithmic}
    \algorule
  \end{minipage}
  \vspace{-\intextsep}
\end{wrapfigure}

% \begin{wrapfigure}[10]{r}{0.5\linewidth} % 14 = approx. lines to reserve; tweak as needed
%   \vspace{-\intextsep}                    % tighten top gap (optional)
%   \begin{minipage}{\linewidth}
%     \captionof{algorithm}{Aerial-Text Vocabulary Expansion}
%     \label{alg:multi_instance_association}
%     \begin{algorithmic}[1]
%         \State \textbf{Initialize:} $D_{\text{text}} \gets \emptyset$ \hfill \textcolor{gray}{\# Empty dataset}
%         \For{$c \in C_{A}$}
%             \State $\mathcal{V}_c \gets \text{ChatGPT}(c)$ \hfill \textcolor{gray}{\# Generate label variations}
%             \State $D_{\text{text}} \gets D_{\text{text}} \cup \{(c, \mathcal{V}_c)\}$
%         \EndFor
%         \State \textbf{Return} $D_{\text{text}}$ \hfill \textcolor{gray}{\# Store for training}
%     \end{algorithmic}
%   \end{minipage}
%   \vspace{-\intextsep}                  
% \end{wrapfigure}

Here, we focus on enhancing category representation by expanding textual variations of aerial object categories. The primary goal is to improve the model’s ability to recognize multiple textual references for the same object, enabling effective aerial object detection in open-vocabulary scenarios.
The process begins by initializing an empty dataset \( D_{\text{text}} \) to store textual variations of aerial object categories. As seen in Algo.~\ref{alg:multi_instance_association}, for each category \( c \) in the aerial dataset \( C_{\text{A}} \), ChatGPT is utilized to generate a set of label variations \( \mathcal{V}_c \). 

\begin{wraptable}{r}{0.5\linewidth}
    \centering
    \vspace{-1em}
    \renewcommand{\arraystretch}{1.0}
    \setlength{\tabcolsep}{4pt}
    \begin{tabular}{p{1.4cm} p{5cm}}
        \toprule
        \textbf{Category} & \textbf{Variations} \\
        \midrule
        \rowcolor{gray!15} Small Aircraft & Light airplane, Private plane, Single-engine aircraft, Propeller plane, Cessna-type aircraft, General aviation aircraft \\
        Helicopter & Chopper, Rotary-wing aircraft, Rotorcraft, Helo, Air ambulance, Military helicopter, Rescue helicopter \\
        \rowcolor{gray!15} Shed & Storage shed, Outbuilding, Toolshed, Garden shed, Small barn \\
        Excavator & Digger, Backhoe, Trackhoe, Mechanical shovel, Hydraulic excavator \\
        \rowcolor{gray!15} Small Car & Compact car, Hatchback, Subcompact vehicle, Economy car, City car, Two-door car, Four-door car \\
        \bottomrule
    \end{tabular}
    \vspace{-0.5em}
    \caption{Vocabulary expansion examples.}
    \vspace{-2em}
    \label{tab:vocab_expansion}
\end{wraptable}

These variations include synonymous terms, alternative phrasings, and domain-specific terminology that may be used to refer to the same object in different contexts. The generated variations are stored in \( D_{\text{text}} \) as pairs of the original category \( c \) and its corresponding set of variations \( \mathcal{V}_c \). Once all categories are processed, the dataset \( D_{\text{text}} \) is finalized and returned for use in training. This expansion follows a hierarchical association where each category serves as a parent node with multiple semantically related variations acting as child nodes. By structuring the data in this way, the model can better interpret and generalize aerial object classes in an open-vocabulary setting. 

It is worth noting that while our data generation pipeline (Aerial-Ground Object Detection Correspondence and Aerial-Text Vocabulary Expansion) provides essential correspondences and variations, the key novelty lies in how these resources are exploited through structured contrastive alignment, rather than in dataset construction itself.

\vspace{-3mm}
\section{Experiments}
\label{sec:experiments}
\vspace{-3mm}

We evaluate zero-shot transfer on unseen aerial benchmarks and compare our approach with state-of-the-art open-vocabulary detectors. We also carry out ablation studies to evaluate the impact of various design decisions, such as encoder patch size, loss contributions etc.

\noindent \textbf{Training Data.} Our pipeline builds cross-view correspondences and expands the vocabulary to strengthen contrastive alignment. Between xView and LVIS, 12 categories overlap and 48 are unique to xView. We create 25,403 positive aerial $\leftrightarrow$ ground pairs via direct matches, rising to 50,650 with inferred correspondences; adding CC12M brings the total to 310,548 pairs. Vocabulary expansion grows 60 aerial categories into 360 variants, broadening textual coverage (examples in Tab.~\ref{tab:vocab_expansion}).

\begin{figure*}[t]
  \centering
  \vspace{2mm}

  \begin{subfigure}[b]{\textwidth}
    \centering
    \includegraphics[width=\textwidth]{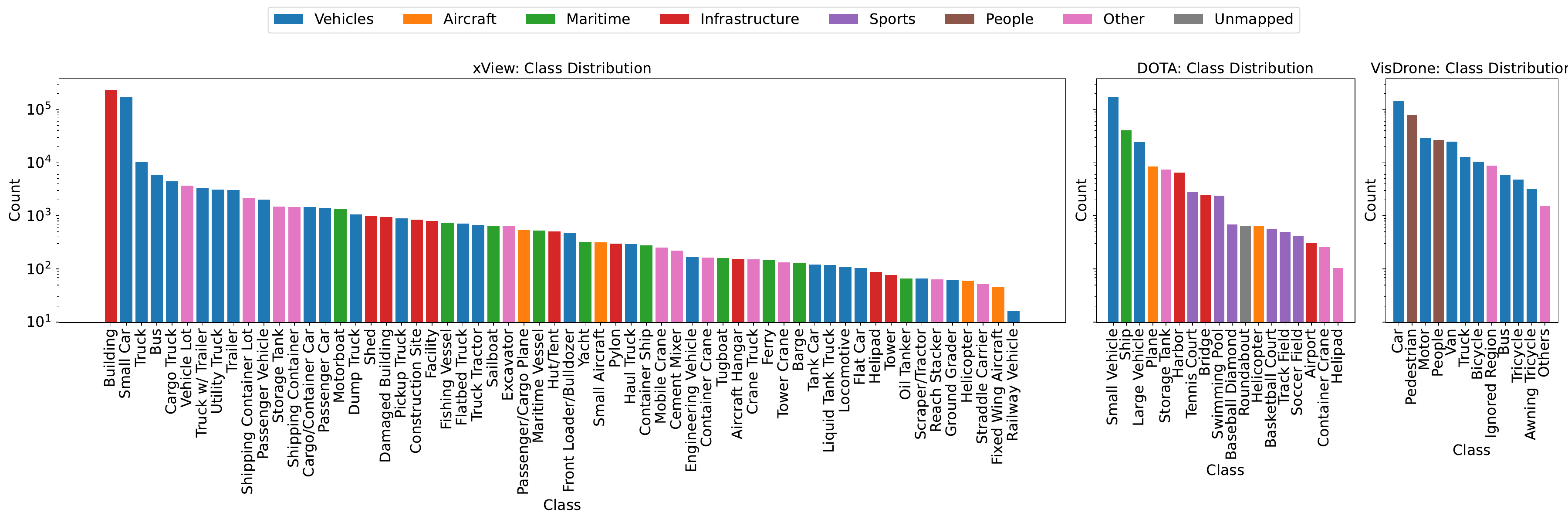}
    \vspace{-3em}
    \caption{\textbf{}}
    \label{fig:subfig3}
  \end{subfigure}
 
  \begin{subfigure}[b]{\textwidth}
    \centering
    \includegraphics[width=\textwidth]{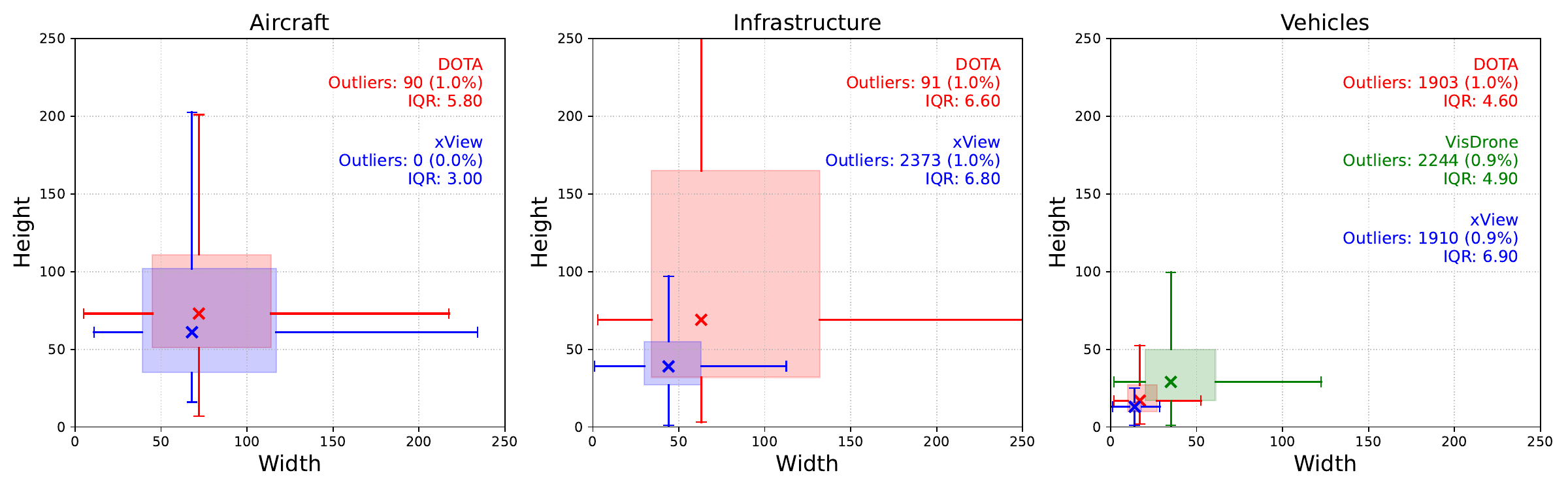}
    \vspace{-2em}
   \caption{\textbf{}}
    \label{fig:subfigsizes}
  \end{subfigure}
  
  \vspace{-0.5em}
  \caption{\textbf{Distribution} of \textbf{(a)} object categories  and \textbf{(b)} size variations across three remote sensing datasets: xView, DOTAv2, and VisDrone (Images).  The figure highlights category imbalance and significant object scale variations, reflecting challenges in open-vocabulary aerial object detection.}
  \label{fig:DatasetComparison}
  \vspace{-3mm}
\end{figure*}

\noindent
\textbf{Evaluation Datasets.} We evaluate generalization on five remote-sensing benchmarks: xView~\cite{lam2018xview}, DOTAv2~\cite{xia2018dota}, VisDrone (images/videos)~\cite{cao2021visdrone}, DIOR~\cite{li2020object}, and HRRSD~\cite{zhang2019hierarchical}. These datasets span satellite, aerial, and drone imagery. Collectively, they contain millions of labeled objects across dozens of categories with wide variation in scale, orientation, and scene complexity, stressing robustness to different sensors, viewpoints, conditions, and object sizes. Fig.~\ref{fig:DatasetComparison} summarizes instance and size distributions, revealing pronounced category imbalance and scale variation, which are key challenges for open-vocabulary aerial object detection.

% \textbf{Training}  
% Our data generation pipeline constructs cross-view correspondences and expands vocabulary to improve contrastive alignment. In the xView-LVIS dataset, 12 object categories are shared between the two datasets, while the remaining 48 categories exist in xView (aerial) but absent in LVIS (ground). We generate 25,403 positive pairs through direct correspondence, increasing to 50,650 with inferred matches. Incorporating CC12M further expands the dataset to 310,548 positive pairs. Additionally, the vocabulary expansion process increases the initial set of 60 aerial categories to 360 distinct textual variations, ensuring a broader textual representation of objects, with some examples shown in Tab. \ref{tab:vocab_expansion}.

\noindent\textbf{Metrics}
For our evaluation, we utilize the standard \( \text{AP}_{50:95} \) metric, adhering to the benchmarking protocol set by MS COCO \cite{lin2014microsoft}. Further to assess overall performance across both seen and unseen object classes we utilize Harmonic Mean (HM) of $\text{mAP}_{\text{base}}$ and $\text{mAP}_{\text{novel}}$ classes.

%This metric calculates the mean Average Precision (AP) across all object categories by averaging detection performance over IoU thresholds from 0.5 to 0.95 in steps of 0.05. By incorporating a wide range of IoU values, it serves as a rigorous and comprehensive measure of detection accuracy, offering deeper insights into the model’s real-world effectiveness.  
\noindent
% Furthermore, we employ the Harmonic Mean (HM) to assess overall performance across both seen and unseen object classes. This ensures a balanced evaluation of the model’s ability to generalize beyond its training data. HM is computed as the harmonic mean of the mean Average Precision (mAP) for base and novel categories, providing a more equitable measure of detection performance across different class distributions.
% \begin{equation}
% \text{HM} = \frac{2 \times \text{mAP}_{\text{base}} \times \text{mAP}_{\text{novel}}}{\text{mAP}_{\text{base}} + \text{mAP}_{\text{novel}}}
% \end{equation}

% \vspace{-3mm}
% \subsection{Quantitative Results}
% \vspace{-3mm}

% \noindent\textbf{Zero-Shot Cross-Dataset Setting}
% Unless stated otherwise, the model is trained only on our generated xView–LVIS cross-view 
% vocabulary-expanded data and evaluated directly on DOTAv2, VisDrone, DIOR, and HRRSD with 
% \emph{no} training data from those test datasets. Consistently, we refer to this protocol as \emph{zero-shot transferability} across the experiments.

\subsection{Zero-Shot Generalization in Aerial Datasets}

\begin{wrapfigure}{r}{0.5\linewidth}
\centering
\vspace{-1em}
\includegraphics[width=\linewidth]{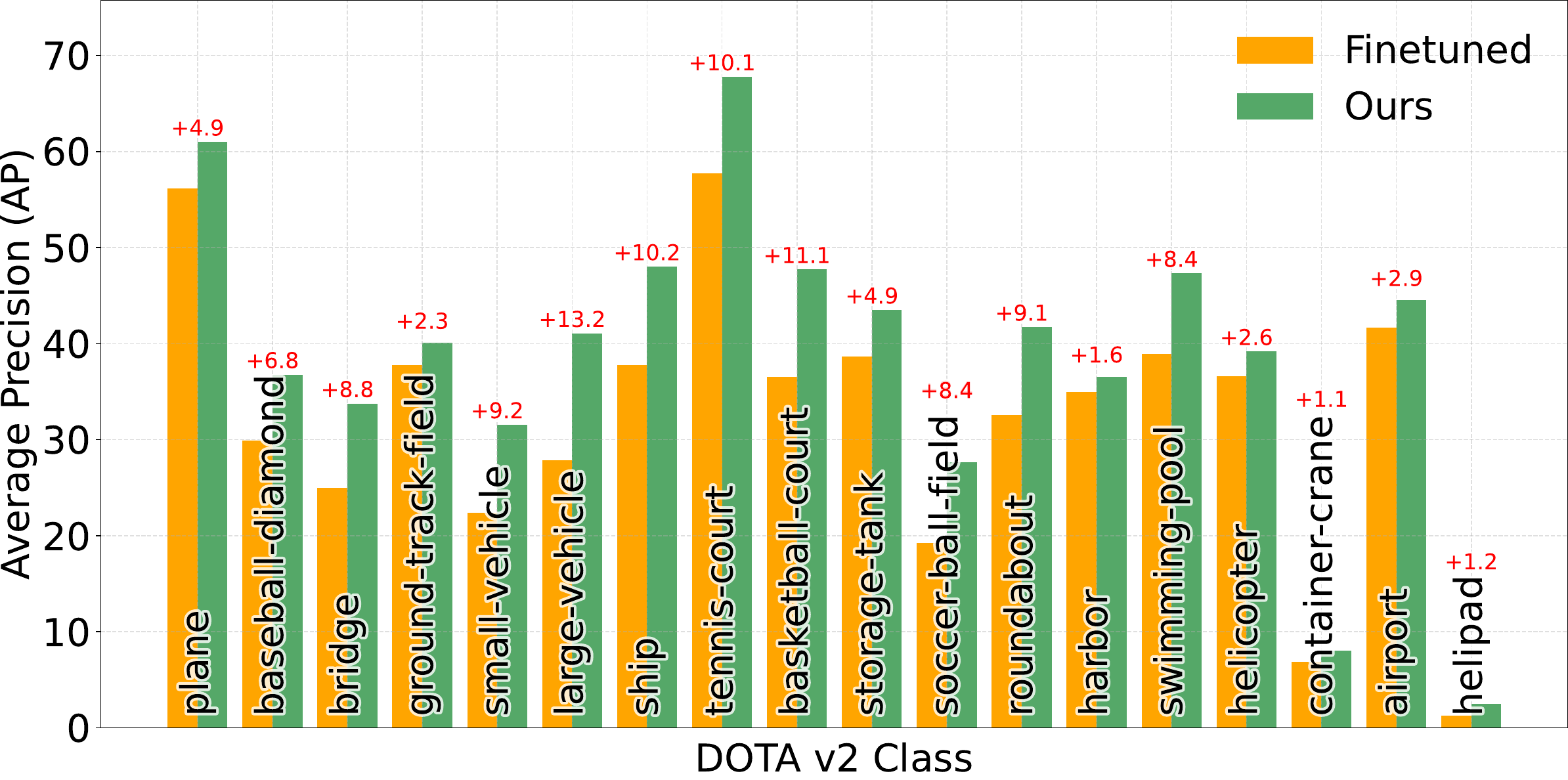}
\vspace{-2em}\caption{Per-class zero-shot performance of our model on DOTAv2 outperforms finetuning.}
% ; 
% for classes without aerial-ground matches (\textcolor{blue}{$\Delta$}), improvements are minimal, 
% aligning with gains from textual association in Tab.~6.}
\label{fig:per_class_performance}
\vspace{-1em}
\end{wrapfigure}

\begin{figure}[h]
  \centering
  \vspace{2mm}
  \includegraphics[width=\linewidth]{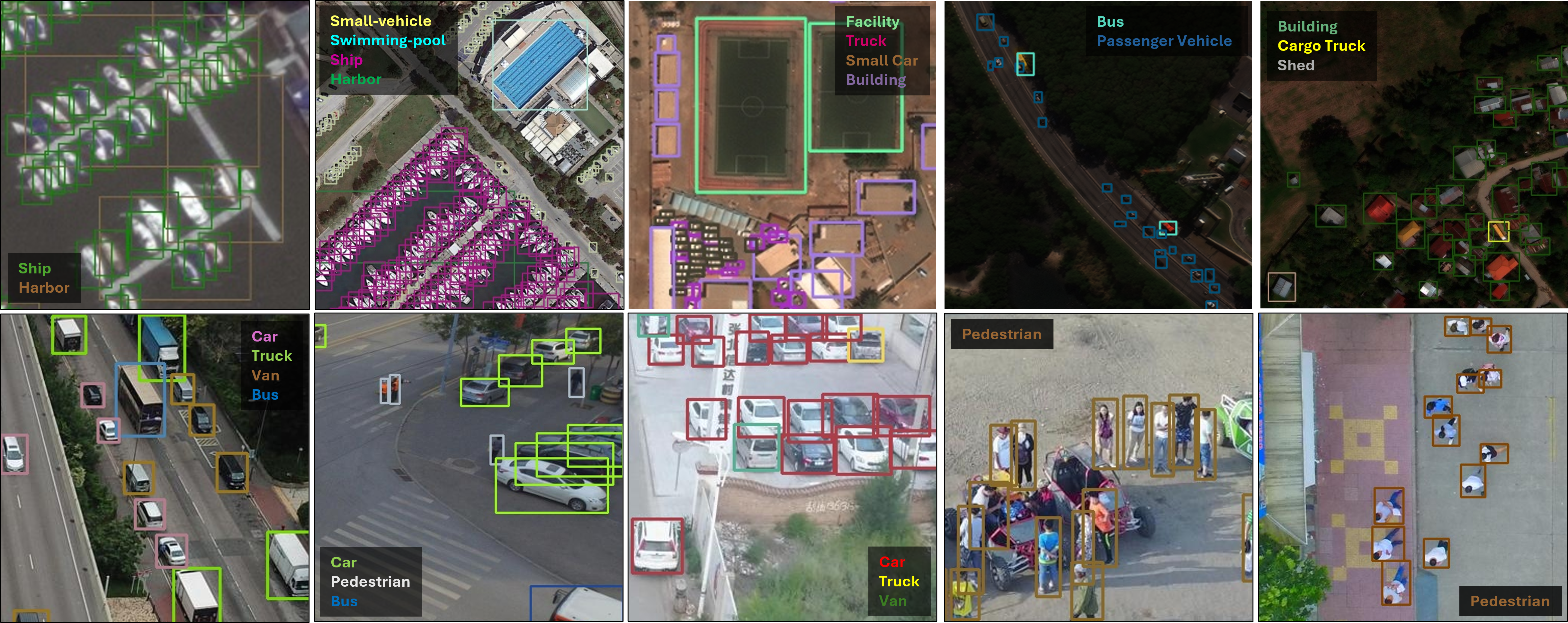}
  \vspace{-3mm}
    % \caption{Qualitative results demonstrating object detection performance across diverse aerial dataset scenarios, including maritime environments, sports facilities, grayscale satellite imagery, urban scenes with large-scale structures, nighttime conditions, and street-level scenarios featuring multiple interacting object classes.}
      \caption{Qualitative results across diverse aerial dataset scenarios, including maritime environments (\textit{1}, \textit{2}), sports facilities (\textit{3}), nighttime conditions (\textit{4}, \textit{5}), urban scenes featuring multiple interacting object classes (\textit{6}, \textit{7}, \textit{8}), and pedestrians across widely varying viewpoints (\textit{9}, \textit{10}).}
  \label{fig:qualitative}
  % \vspace{-3mm}
\end{figure}

Our model is trained only on our generated xView–LVIS cross-view vocabulary-expanded data and evaluated in a zero-shot manner on DOTAv2, VisDrone, DIOR, and HRRSD, with 
\emph{no} training data from those test datasets. In all experiments, we consistently refer to this protocol as \emph{zero-shot transferability}. In Tab.~\ref{tab:zero-shot}, we compare our zero-shot object detection approach against YOLOv11 (finetuned) \cite{yolo11_ultralytics}, which is trained on the entire dataset, across six aerial datasets: xView, DOTAv2, VisDrone (Images), VisDrone (Videos), DIOR, and HRRSD. In contrast, our model never sees these datasets during training, yet achieves consistent improvements in a zero-shot setting, with performance gains of \textbf{+6.32} on DOTAv2, \textbf{+4.16} on VisDrone (Images), \textbf{+2.37} on VisDrone (Videos), \textbf{+2.23} on DIOR, and \textbf{+3.46} on HRRSD over YOLO. For xView, which is not evaluated in a true zero-shot setting (*), our approach still improves by \textbf{+2.99}. These results highlight our model’s strong cross-domain generalization, demonstrating competitive performance without dataset-specific finetuning. Refer to Fig.~\ref{fig:qualitative} for qualitative results across datasets and Fig.~\ref{fig:per_class_performance} for the per-class breakdown on DOTAv2.

\vspace{-3mm}
\begin{table}[h]
    \centering
    \renewcommand{\arraystretch}{1.0}
    \setlength{\tabcolsep}{3pt}
    \begin{tabular}{lccc}
        \toprule
        \textbf{Datasets} & \multicolumn{2}{c}{\textbf{Finetuned}} & \textbf{Zero-Shot}\\ 
        \cmidrule(lr){2-3}
        % \multicolumn{2}{r}{\textbf{Finetuned on Target Dataset}} & \textbf{Zero-Shot} \\
        & \textit{OWLv2} & \textit{YOLOv11} & \textit{Ours} \\
        \midrule
        \rowcolor{gray!15} xView \cite{lam2018xview} & 12.65 & 34.92 & 37.91$^{*}$~\increase{(+2.99)} \\  
        DOTAv2 \cite{xia2018dota} & 15.01 & 32.28 & 38.60~\increase{(+6.32)} \\
        \rowcolor{gray!15} VisDrone (Images) \cite{cao2021visdrone} & 25.70 & 40.81 & 44.97~\increase{(+4.16)} \\
        VisDrone (Videos) \cite{cao2021visdrone} & 18.44 & 34.65 & 37.02~\increase{(+2.37)} \\
        \rowcolor{gray!15} DIOR \cite{li2020object} & 45.91 & 61.68 & 63.91~\increase{(+2.23)} \\
        HRRSD \cite{zhang2019hierarchical} & 53.25 & 70.66 & 74.12~\increase{(+3.46)} \\
        \bottomrule
    \end{tabular}
    \caption{Comparison of OWLv2 (Finetuned), YOLO (Finetuned) and our zero-shot approach across multiple datasets. Note: `*' xView is not evaluated in a zero-shot setting.}
    \label{tab:zero-shot}
    \vspace{-2mm}
\end{table}

% \begin{figure}
%   \centering
%   \vspace{2mm}
%   \includegraphics[width=0.7\linewidth]{images/Qualitative.png}
%   \vspace{-3mm}
%     \caption{Qualitative results demonstrating object detection performance across diverse aerial dataset scenarios, including maritime environments, sports facilities, grayscale satellite imagery, urban scenes with large-scale structures, nighttime conditions, and street-level scenarios featuring multiple interacting object classes.}
%   \label{fig:qualitative}
%   \vspace{-3mm}
% \end{figure} 

% \begin{figure}
%   \centering
%   \vspace{2mm}
%   \includegraphics[width=\linewidth]{images/Qualitative_plus_PerClass.png}
%   \vspace{-3mm}
%     % \caption{Qualitative results demonstrating object detection performance across diverse aerial dataset scenarios, including maritime environments, sports facilities, grayscale satellite imagery, urban scenes with large-scale structures, nighttime conditions, and street-level scenarios featuring multiple interacting object classes.}
%       \caption{
%   \textbf{Performance} 
%   Qualitative results demonstrating object detection performance across diverse aerial dataset scenes, including maritime environments (\textit{1}, \textit{2}), sports facilities (\textit{3}), nighttime conditions (\textit{4}, \textit{5}), urban scenes (\textit{6}), and pedestrians across different viewpoints (\textit{7}, \textit{8}). Also, per-class performance on DOTAv2 showing consistent gains over simple finetuning (\textit{9}). }
%   \label{fig:qualitative}
%   \vspace{-3mm}
% \end{figure} 

\begin{table*}[h]
    \centering
    \renewcommand{\arraystretch}{0.8}
    \setlength{\tabcolsep}{6pt}
    \begin{tabular}{l c c c c c c c}
        \toprule
        \textbf{Method} & \textbf{External} & $\mathcal{T}_{novel}$ & $\mathcal{T}_{cls}$ & \textbf{mAP} & \textbf{mAP$_{base}$} & \textbf{mAP$_{novel}$} & \textbf{HM} \\
        \midrule
        \rowcolor{gray!15} Detic & NWPU-RESISC45 & \ding{55} & \ding{51} & 16.8 & 19.8 & 4.8 & 7.7 \\
        ViLD & DIOR & \ding{51} & \ding{55} & 25.6 & 28.5 & 14.2 & 19.0 \\
        \rowcolor{gray!15} OV-DETR & DIOR & \ding{51} & \ding{55} & 25.6 & 25.6 & 25.5 & 25.6 \\
        BARON & NWPU-RESISC45 & \ding{55} & \ding{51} & 27.4 & 29.4 & 19.5 & 23.5 \\
        \rowcolor{gray!15} YOLO-World & NWPU-RESISC45 & \ding{55} & \ding{51} & 32.9 & 39.1 & 8.5 & 13.9 \\
        GroundingDINO & NWPU-RESISC45 & \ding{55} & \ding{51} & 33.0 & 40.5 & 3.3 & 6.1 \\
        \rowcolor{gray!15} GLIP & NWPU-RESISC45 & \ding{55} & \ding{51} & 33.8 & \secondbest{41.0} & 5.4 & 9.5 \\
        CastDet & DIOR & \ding{55} & \ding{55} & 38.5 & 36.5 & \secondbest{46.5} & 40.9 \\
        \rowcolor{gray!15} CastDet & DIOR & \ding{51} & \ding{55} & \secondbest{40.5} & 39.0 & 46.3 & \secondbest{42.3} \\
        \midrule
        \textit{Ours} & xView & \ding{55} & \ding{55} & \textcolor{red}{\bf 44.9} & \textcolor{red}{\bf 42.7} & \textcolor{red}{\bf 49.2} & \textcolor{red}{\bf45.7} \\
        \bottomrule
    \end{tabular}
    \vspace{-0.5em}
    \caption{Performance comparison of SOTA open-vocabulary object detection methods on the VisDrone (Images) dataset. Here, \(\mathcal{T}_{novel}\) indicates whether novel classes are pre-known, while \(\mathcal{T}_{cls}\) denotes whether additional classification or caption datasets are used during training. The best method is shown in \bestresult{Red}, and the second-best method is shown in \secondbest{Blue}.}
    \vspace{-0.5em}
    \label{tab:SOTA}
\end{table*}
\vspace{-3mm}
\subsection{Comparison to State-of-the-Art}
\vspace{-3mm}
% Tab.~\ref{tab:SOTA} further strengthens this finding by demonstrating that our method surpasses existing state-of-the-art open-vocabulary object detection approaches on the VisDrone (Images) dataset. 

As seen in Tab.~\ref{tab:SOTA}, our method surpasses existing state-of-the-art open-vocabulary object detection approaches on the VisDrone (Images) dataset. These gains underscore the effectiveness of our aerial-ground alignment and aerial-text association.

\vspace{-3mm}
\subsection{Ablation Studies}
\subsubsection{Vision Encoder Patch Size}
\begin{wraptable}{r}{0.5\linewidth}
    \centering
    \vspace{-2em}
    \renewcommand{\arraystretch}{1.0}
    \setlength{\tabcolsep}{1pt}
    \begin{tabular}{lcc}
        \toprule
         \textbf{Datasets} & \multicolumn{2}{c}{\textbf{Patch Size}} \\
         \cmidrule(lr){2-3}
        & \textit{Ours (/16)} & \textit{Ours (/14)} \\
        \midrule
        \rowcolor{gray!15} xView  & 36.48$^{*}$ & 37.91$^{*}$~\increase{(+1.43)} \\
        DOTAv2 & 33.97 & 38.60~\increase{(+4.63)} \\
        \rowcolor{gray!15} VisDrone (Images)  & 42.66 & 44.97~\increase{(+2.31)} \\
        VisDrone (Videos)   & 35.78 & 37.02~\increase{(+1.24)} \\
        \bottomrule
    \end{tabular}
    \vspace{-0.75em}
    \caption{Impact of encoder patch size on open-vocabulary aerial detection performance. Note: `*' indicates xView is not evaluated zero-shot.}
    \label{tab:patch-size}
    \vspace{-1em}
\end{wraptable}

% \begin{table}[h]
%     \centering
%     \renewcommand{\arraystretch}{1.0}
%     \setlength{\tabcolsep}{2pt}
%     \begin{tabular}{lcc}
%         \toprule
%         \textbf{Datasets} & \textbf{Ours (Patch-16)} & \textbf{Ours (Patch-14)} \\
%         \midrule
%         \rowcolor{gray!15} xView \cite{lam2018xview} & 36.48$^{*}$ & 37.91$^{*}$~\increase{(+1.43)} \\
%         DOTAv2 \cite{xia2018dota} & 33.97 & 38.60~\increase{(+4.63)} \\
%         \rowcolor{gray!15} VisDrone (Images) \cite{cao2021visdrone} & 42.66 & 44.97~\increase{(+2.31)} \\
%         VisDrone (Videos) \cite{cao2021visdrone}  & 35.78 & 37.02~\increase{(+1.24)} \\
%         \bottomrule
%     \end{tabular}
%     \vspace{-0.75em}
%     \caption{Ablation study evaluating the impact of encoder patch size on open-vocabulary aerial detection performance. Note: `*' indicates that xView is not evaluated in a zero-shot setting, as it is included in our cross-view data.}
%     \label{tab:patch-size}
% \end{table}
The ablation study in Tab.~\ref{tab:patch-size} assesses the impact of vision encoder patch size (Patch-16 vs. Patch-14) on open-vocabulary aerial object detection. Smaller patch sizes (Patch-14) consistently improve performance across all datasets, with DOTAv2 showing the largest gain of \textbf{+4.63}. This improvement is likely due to Patch-14 capturing finer spatial details, leading to better feature granularity and enhanced object discrimination, especially in high-variance aerial imagery. 
\subsubsection{Aerial-Ground Correspondence Dataset}

% \begin{table}[h]
%     \centering
%     \renewcommand{\arraystretch}{1.0}
%     \setlength{\tabcolsep}{8pt}
%     \begin{tabular}{lcc}
%         \toprule
%         \multirow{2}{*}{\textbf{Evaluation Datasets}} & 
%         \multicolumn{2}{c}{\textbf{Aerial-Ground}} \\
%          \multirow{2}{*}{\textbf{}} & 
%         \multicolumn{2}{c}{\textbf{Correspondence Datasets}} \\
%         \cmidrule(lr){2-3}
%         & \textit{LVIS} & \textit{CC12M} \\
%         \midrule
%         \rowcolor{gray!15} xView \cite{lam2018xview}  & \multicolumn{1}{c}{33.25} & \multicolumn{1}{c}{36.48} \\
%         DOTAv2 \cite{xia2018dota}  & \multicolumn{1}{c}{27.10} & \multicolumn{1}{c}{33.97} \\
%         \bottomrule
%     \end{tabular}
%     \vspace{-0.5em}
%     \caption{Performance comparison on xView and DOTAv2 using different aerial-ground correspondence datasets (LVIS, CC12M) to evaluate the impact of alignment dataset size on cross-view representation learning.}
%     \label{tab:aerial_ground_corr}
%     \vspace{-0.5em}
% \end{table}
\begin{wraptable}{r}{0.4\linewidth}
    \centering
    \vspace{-2em}
    \renewcommand{\arraystretch}{1.0}
    \setlength{\tabcolsep}{6pt}
    \begin{tabular}{lcc}
        \toprule
        \textbf{Evaluation} & 
        \multicolumn{2}{c}{\textbf{Aerial-Ground}} \\
         \textbf{Dataset} & 
        \multicolumn{2}{c}{\textbf{Correspondences}} \\
        \cmidrule(lr){2-3}
        & \textit{LVIS} & \textit{CC12M} \\
        \midrule
        \rowcolor{gray!15} xView  & \multicolumn{1}{c}{33.25} & \multicolumn{1}{c}{36.48} \\
        DOTAv2  & \multicolumn{1}{c}{27.10} & \multicolumn{1}{c}{33.97} \\
        \bottomrule
    \end{tabular}
    \vspace{-0.5em}
    \caption{Performance comparison on xView and DOTAv2 using different datasets (LVIS, CC12M) as the source of aerial-ground correspondence.}
    \label{tab:aerial_ground_corr}
    \vspace{-1em}
\end{wraptable}
Tab.~\ref{tab:aerial_ground_corr} compares detection performance on xView and DOTAv2 when using different aerial-ground correspondence datasets (LVIS vs. CC12M) for alignment. CC12M leads to a substantial performance improvement across both datasets (\textbf{+3.23} on xView, \textbf{+6.87} on DOTAv2), highlighting the benefits of a larger and more diverse dataset for cross-view feature alignment. Aerial-ground alignment benefits from larger and more diverse datasets, as seen with CC12M outperforming LVIS. This suggests that training with a broader distribution of images improves cross-view generalization and enhances feature transferability, particularly for open-vocabulary aerial detection. The effect is more pronounced on DOTAv2, due to its higher scene complexity, where additional aerial-ground associations provide stronger supervision.

\subsubsection{Contribution of Contrastive Losses}

\begin{wraptable}{r}{0.45\textwidth}
  \centering
  \renewcommand{\arraystretch}{1.0}
  \setlength{\tabcolsep}{1pt}
  \vspace{-2em}
  \begin{tabular}{lccc}
    \toprule
    \textbf{Datasets} & \multicolumn{3}{c}{\textbf{Contrastive Losses}} \\
    \cmidrule(lr){2-4}
    \textbf{} & \textit{\textcolor{darkyellow}{Image\textsubscript{A}}--\textcolor{cadmiumgreen}{Text}} & 
      \textit{\textcolor{darkyellow}{Image\textsubscript{A}}--\textcolor{orange}{Image\textsubscript{G}}} & \textit{Both} \\
    \midrule
    \rowcolor{gray!15} xView & 26.13 & 32.41 & 33.25 \\
    DOTAv2  & 19.98 & 25.06 & 27.10 \\
    \bottomrule
  \end{tabular}
  \vspace{-0.5em}
  \caption{Impact of individual losses.}
  \vspace{-1em}
  \label{tab:contrastive_losses}
\end{wraptable}

% \begin{table}[h]
%     \centering
%     \renewcommand{\arraystretch}{1.0}
%     \setlength{\tabcolsep}{4pt}
%     \begin{tabular}{l c{2.5cm} c{2.5cm} c{2.5cm}}
%         \toprule
%         \textbf{Datasets} & \multicolumn{3}{c}{\textbf{Contrastive Losses}} \\
%         \cmidrule(lr){2-4}
%         & \textit{\textcolor{cadmiumgreen}{Image\textsubscript{A}}--\textcolor{purple}{Text}} & 
%           \textit{\textcolor{cadmiumgreen}{Image\textsubscript{A}}--\textcolor{blue}{Image\textsubscript{G}}} & \textit{Cumulative} \\
%         \midrule
%         \rowcolor{gray!15} xView \cite{lam2018xview}  & 26.13 & 32.41 & 33.25 \\
%         DOTAv2 \cite{xia2018dota}  & 19.98 & 25.06 & 27.10 \\
%         \bottomrule
%     \end{tabular}
%     \vspace{-0.5em}
%     \caption{Impact of individual contrastive losses on cumulative performance for xView and DOTAv2.}
%     \label{tab:contrastive_losses}
% \end{table}

The individual contributions of \textit{\textcolor{darkyellow}{Image\textsubscript{A}}--\textcolor{cadmiumgreen}{Text}} and \textit{\textcolor{darkyellow}{Image\textsubscript{A}}--\textcolor{orange}{Image\textsubscript{G}}} contrastive losses to overall cumulative performance on the xView and DOTAv2 datasets are evaluated in Tab.~\ref{tab:contrastive_losses}. The \textit{\textcolor{darkyellow}{Image\textsubscript{A}}--\textcolor{cadmiumgreen}{Text}} contrastive loss aligns aerial image features with textual representations, facilitating open-vocabulary recognition, whereas the \textit{\textcolor{darkyellow}{Image\textsubscript{A}}--\textcolor{orange}{Image\textsubscript{G}}} contrastive loss enforces aerial-to-ground view consistency, improving domain adaptation. The cumulative score reflects the combined effect of both losses, highlighting that \textit{\textcolor{darkyellow}{Image\textsubscript{A}}--\textcolor{orange}{Image\textsubscript{G}}} contrastive learning plays a more dominant role in feature alignment, as indicated by its higher individual contribution. However, the synergy between both losses leads to the best overall performance, demonstrating the importance of multi-modal representation learning in cross-view object detection.

\vspace{-3mm}
\section{Conclusion}
\vspace{-3mm}
We propose a novel cross-view contrastive alignment framework for open-vocabulary object detection in aerial imagery. To facilitate this, we establish aerial-ground correspondences using xView, LVIS, and CC12M, generating a large corpus of correspondence pairs. To bridge the modality gap between aerial and ground views, we establish cross-view alignment by learning semantic correspondences between detected objects across perspectives. Furthermore, we enhance generalization by employing multi-instance contrastive learning to align aerial images with their textual descriptions. Through extensive experiments, we demonstrate that our approach in a zero-shot setting outperforms finetuned models, advancing the scalability and adaptability of aerial object detection.

\noindent
\textbf{Acknowledgment} \\
This work was conducted at UCF and supported by Lockheed Martin
Corporate Engineering.

\newpage
\bibliography{iclr2026_conference}
\bibliographystyle{iclr2026_conference}

\newpage
\appendix
\section{Appendix}
\vspace{-3mm}
\subsection{Model-Agnostic Contrastive Alignment Strategy}
\vspace{-3mm}
Experimental results on the VisDrone dataset demonstrate the model-agnostic nature of our framework, as shown using OWLv2 and GroundingDINO. Given any open-vocabulary ground-view detector, we generate cross-view data which, combined with our contrastive alignment strategy, enables the construction of an open-vocabulary aerial-view detector—without being tied to a specific architecture like OWLv2. As seen in \ref{tab:model_agnostic}, the framework transfers seamlessly across models and facilitates effective ground-to-aerial adaptation.

\begin{table}[h]
\centering
\renewcommand{\arraystretch}{0.98}
\setlength{\tabcolsep}{2pt}
\begin{tabular}{lcc}
\toprule
\textbf{Model} & \textbf{Finetuned} & \textbf{Ours} \textit{mAP} \\
\midrule
OWLv2 \cite{minderer2023scaling} & 25.70 & 44.97~\increase{(+19.27)} \\
\rowcolor{gray!15} GroundingDINO \cite{liu2024grounding} & 33.00 & 47.15~\increase{(+14.15)} \\
\bottomrule
\end{tabular}
\caption{Model-agnostic results on the VisDrone dataset.}
\label{tab:model_agnostic}
\end{table}

\vspace{-3mm}
\subsection{Qualitative Results}
\vspace{-3mm}
\begin{figure*}[h]
  \centering
  \includegraphics[width=1\textwidth]{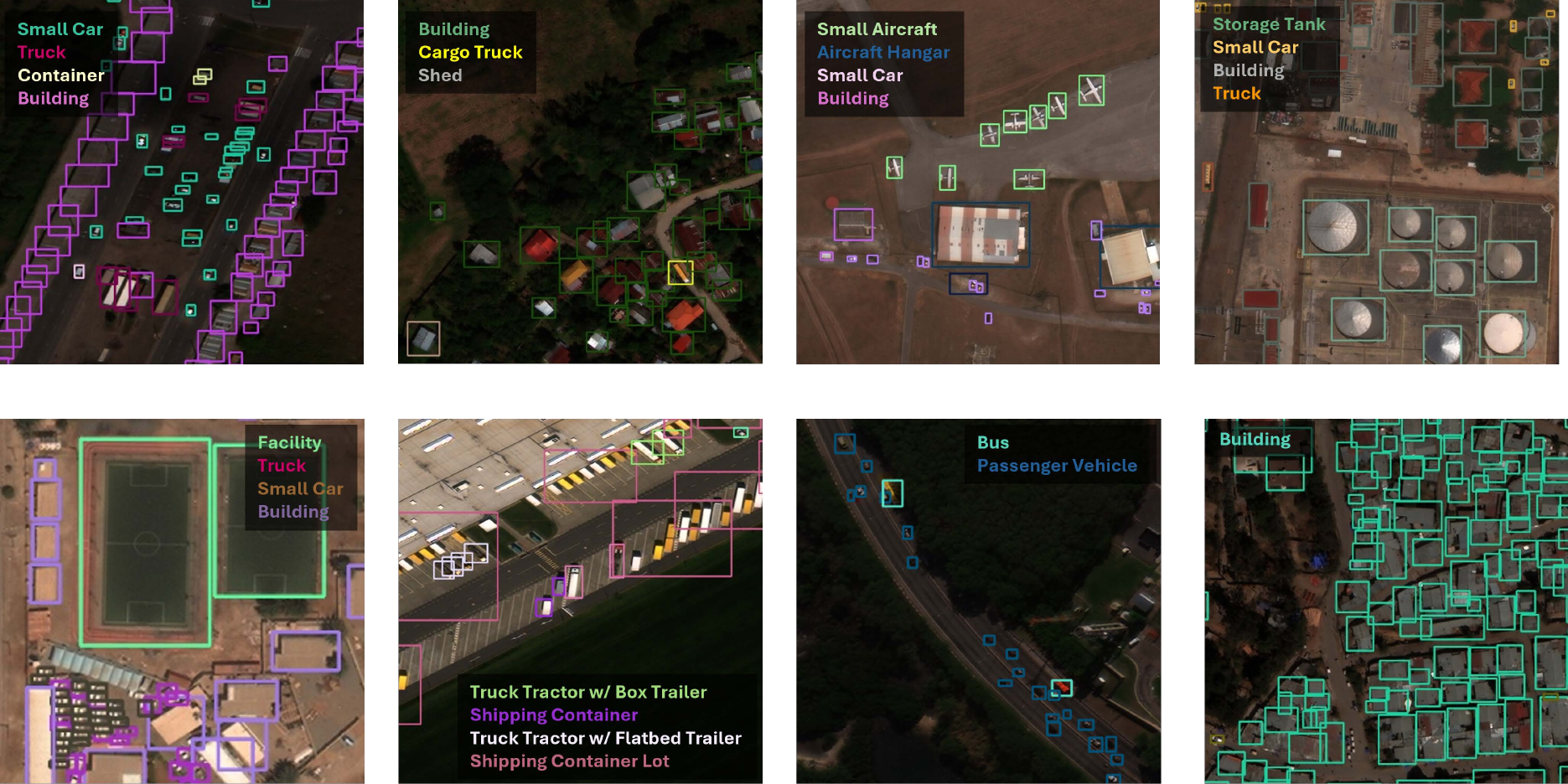}
  \vspace{-4mm}
  \caption{Qualitative results on xView \cite{lam2018xview}}
  \label{fig:Appendix_xView}
  \vspace{-4mm}
\end{figure*}

% \vspace{-3mm}
\begin{figure*}[h]
  \centering
  \includegraphics[width=1\textwidth]{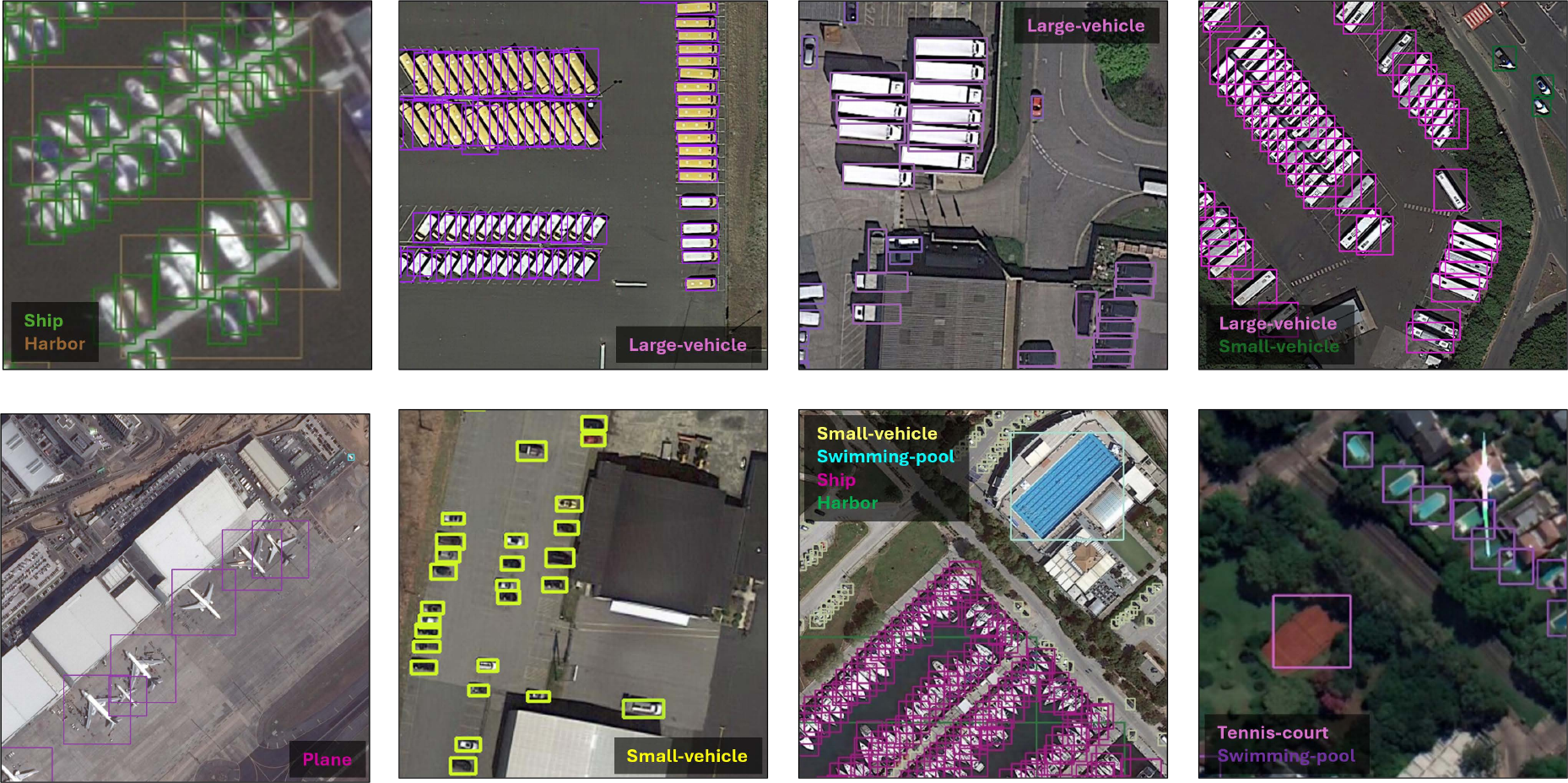}
  \vspace{-4mm}
  \caption{Qualitative results on DOTAv2 \cite{xia2018dota}}
  \label{fig:Appendix_DOTA}
  \vspace{-3mm}
\end{figure*}

\clearpage

\begin{figure}[H]
  \centering
  \vspace{2mm}
  \includegraphics[width=1\textwidth]{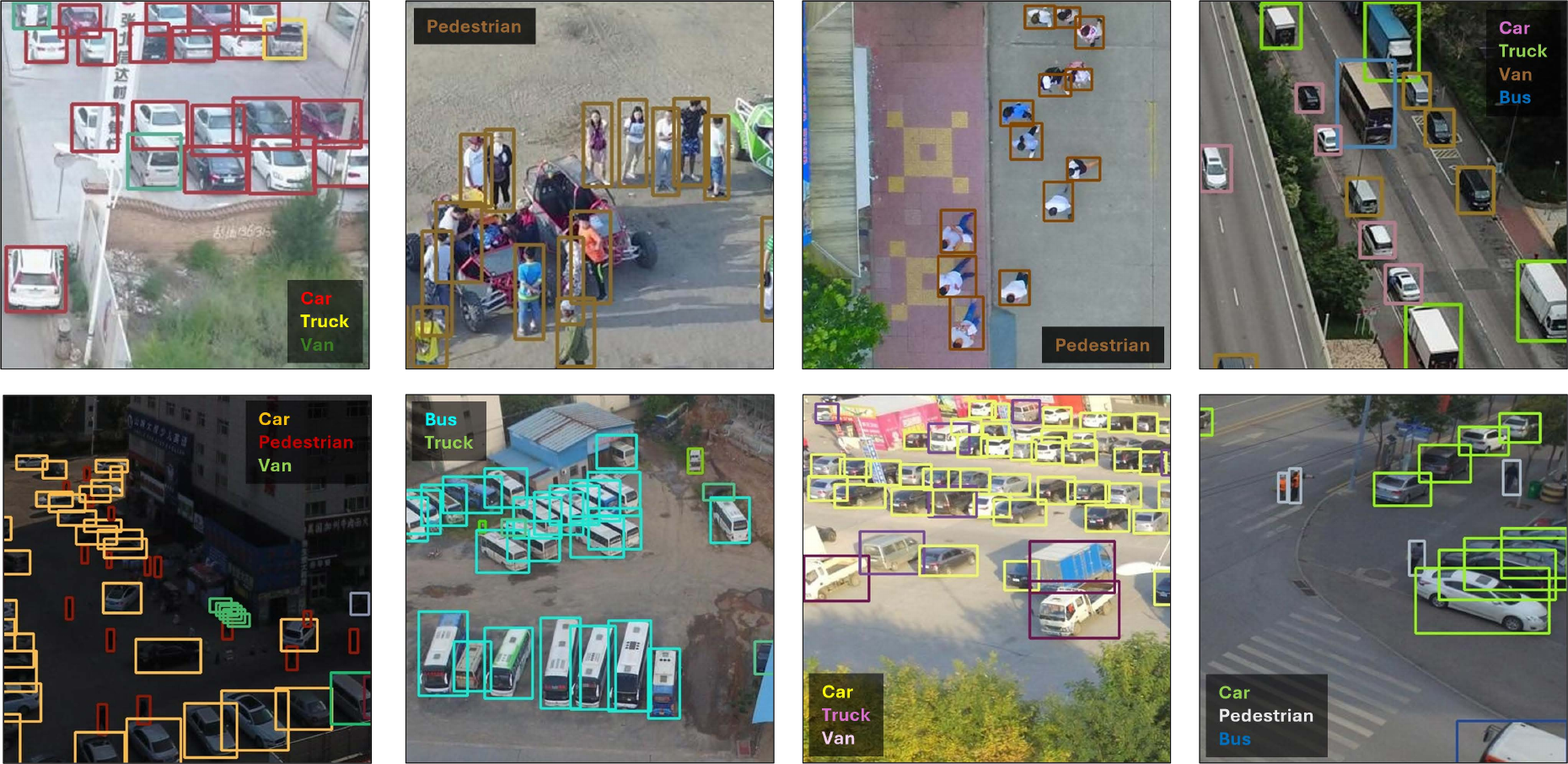}
  \vspace{-4mm}
  \caption{Qualitative results on VisDrone \cite{cao2021visdrone}}
  \label{fig:Appendix_VisDrone}
  \vspace{-3mm}
\end{figure}

\end{document}

%% file: math_commands.tex
\usepackage{amsmath,amsfonts,bm}

\def\eqref#1{equation~\ref{#1}}
\def\1{\bm{1}}

\DeclareMathAlphabet{\mathsfit}{\encodingdefault}{\sfdefault}{m}{sl}
\SetMathAlphabet{\mathsfit}{bold}{\encodingdefault}{\sfdefault}{bx}{n}

%% file: root.bbl
\begin{thebibliography}{59}
\providecommand{\natexlab}[1]{#1}
\providecommand{\url}[1]{\texttt{#1}}
\expandafter\ifx\csname urlstyle\endcsname\relax
  \providecommand{\doi}[1]{doi: #1}\else
  \providecommand{\doi}{doi: \begingroup \urlstyle{rm}\Url}\fi

\bibitem[Ak et~al.(2024)Ak, Mohta, Dimitriadis, Manchanda, Xu, and Shen]{ak2024aligning}
Kenan~Emir Ak, Jay Mohta, Dimitris Dimitriadis, Saurav Manchanda, Yan Xu, and Mingwei Shen.
\newblock Aligning vision language models with contrastive learning.
\newblock 2024.

\bibitem[Cao et~al.(2021)Cao, He, Wang, Wang, Yuan, Zhang, Zhang, Zhu, Van~Gool, Han, et~al.]{cao2021visdrone}
Yaru Cao, Zhijian He, Lujia Wang, Wenguan Wang, Yixuan Yuan, Dingwen Zhang, Jinglin Zhang, Pengfei Zhu, Luc Van~Gool, Junwei Han, et~al.
\newblock Visdrone-det2021: The vision meets drone object detection challenge results.
\newblock In \emph{Proceedings of the IEEE/CVF International conference on computer vision}, pp.\  2847--2854, 2021.

\bibitem[Carion et~al.(2020)Carion, Massa, Synnaeve, Usunier, Kirillov, and Zagoruyko]{carion2020end}
Nicolas Carion, Francisco Massa, Gabriel Synnaeve, Nicolas Usunier, Alexander Kirillov, and Sergey Zagoruyko.
\newblock End-to-end object detection with transformers.
\newblock In \emph{European conference on computer vision}, pp.\  213--229. Springer, 2020.

\bibitem[Changpinyo et~al.(2021)Changpinyo, Sharma, Ding, and Soricut]{changpinyo2021conceptual}
Soravit Changpinyo, Piyush Sharma, Nan Ding, and Radu Soricut.
\newblock Conceptual 12m: Pushing web-scale image-text pre-training to recognize long-tail visual concepts.
\newblock In \emph{Proceedings of the IEEE/CVF conference on computer vision and pattern recognition}, pp.\  3558--3568, 2021.

\bibitem[Chen et~al.(2021)Chen, Savakis, Diehl, Blasch, Wei, and Chen]{chen2021targeted}
Hua-Mei Chen, Andreas Savakis, Ashley Diehl, Erik Blasch, Sixiao Wei, and Genshe Chen.
\newblock Targeted adversarial discriminative domain adaptation.
\newblock \emph{Journal of Applied Remote Sensing}, 15\penalty0 (3):\penalty0 038504--038504, 2021.

\bibitem[Cheng et~al.(2024)Cheng, Song, Ge, Liu, Wang, and Shan]{cheng2024yolo}
Tianheng Cheng, Lin Song, Yixiao Ge, Wenyu Liu, Xinggang Wang, and Ying Shan.
\newblock Yolo-world: Real-time open-vocabulary object detection.
\newblock In \emph{Proceedings of the IEEE/CVF Conference on Computer Vision and Pattern Recognition}, pp.\  16901--16911, 2024.

\bibitem[Cui et~al.(2022)Cui, Zhou, Guo, Yin, Wu, Yoshie, and Chen]{cui2022contrastive}
Quan Cui, Boyan Zhou, Yu~Guo, Weidong Yin, Hao Wu, Osamu Yoshie, and Yubo Chen.
\newblock Contrastive vision-language pre-training with limited resources.
\newblock In \emph{European Conference on Computer Vision}, pp.\  236--253. Springer, 2022.

\bibitem[Ding et~al.(2021)Ding, Xue, Xia, Bai, Yang, Yang, Belongie, Luo, Datcu, Pelillo, et~al.]{xia2018dota}
Jian Ding, Nan Xue, Gui-Song Xia, Xiang Bai, Wen Yang, Michael~Ying Yang, Serge Belongie, Jiebo Luo, Mihai Datcu, Marcello Pelillo, et~al.
\newblock Object detection in aerial images: A large-scale benchmark and challenges.
\newblock \emph{IEEE transactions on pattern analysis and machine intelligence}, 44\penalty0 (11):\penalty0 7778--7796, 2021.

\bibitem[Du et~al.(2022)Du, Wei, Zhang, Shi, Gao, and Li]{du2022learning}
Yu~Du, Fangyun Wei, Zihe Zhang, Miaojing Shi, Yue Gao, and Guoqi Li.
\newblock Learning to prompt for open-vocabulary object detection with vision-language model.
\newblock In \emph{Proceedings of the IEEE/CVF conference on computer vision and pattern recognition}, pp.\  14084--14093, 2022.

\bibitem[Feng et~al.(2022)Feng, Zhong, Jie, Chu, Ren, Wei, Xie, and Ma]{feng2022promptdet}
Chengjian Feng, Yujie Zhong, Zequn Jie, Xiangxiang Chu, Haibing Ren, Xiaolin Wei, Weidi Xie, and Lin Ma.
\newblock Promptdet: Towards open-vocabulary detection using uncurated images.
\newblock In \emph{European conference on computer vision}, pp.\  701--717. Springer, 2022.

\bibitem[Girshick(2015)]{girshick2015fast}
Ross Girshick.
\newblock Fast r-cnn.
\newblock In \emph{Proceedings of the IEEE international conference on computer vision}, pp.\  1440--1448, 2015.

\bibitem[Gu et~al.(2021)Gu, Lin, Kuo, and Cui]{gu2021open}
Xiuye Gu, Tsung-Yi Lin, Weicheng Kuo, and Yin Cui.
\newblock Open-vocabulary object detection via vision and language knowledge distillation.
\newblock \emph{arXiv preprint arXiv:2104.13921}, 2021.

\bibitem[Gupta et~al.(2019)Gupta, Dollar, and Girshick]{gupta2019lvis}
Agrim Gupta, Piotr Dollar, and Ross Girshick.
\newblock Lvis: A dataset for large vocabulary instance segmentation.
\newblock In \emph{Proceedings of the IEEE/CVF conference on computer vision and pattern recognition}, pp.\  5356--5364, 2019.

\bibitem[He et~al.(2017)He, Gkioxari, Doll{\'a}r, and Girshick]{he2017mask}
Kaiming He, Georgia Gkioxari, Piotr Doll{\'a}r, and Ross Girshick.
\newblock Mask r-cnn.
\newblock In \emph{Proceedings of the IEEE international conference on computer vision}, pp.\  2961--2969, 2017.

\bibitem[Hou et~al.(2023)Hou, Lu, Guo, Liu, Gong, Zhu, and Ping]{hou2023feature}
Qingfeng Hou, Jun Lu, Haitao Guo, Xiangyun Liu, Zhihui Gong, Kun Zhu, and Yifan Ping.
\newblock Feature relation guided cross-view image based geo-localization.
\newblock \emph{Remote Sensing}, 15\penalty0 (20):\penalty0 5029, 2023.

\bibitem[Jia et~al.(2021)Jia, Yang, Xia, Chen, Parekh, Pham, Le, Sung, Li, and Duerig]{jia2021scaling}
Chao Jia, Yinfei Yang, Ye~Xia, Yi-Ting Chen, Zarana Parekh, Hieu Pham, Quoc Le, Yun-Hsuan Sung, Zhen Li, and Tom Duerig.
\newblock Scaling up visual and vision-language representation learning with noisy text supervision.
\newblock In \emph{International conference on machine learning}, pp.\  4904--4916. PMLR, 2021.

\bibitem[Jocher \& Qiu(2024)Jocher and Qiu]{yolo11_ultralytics}
Glenn Jocher and Jing Qiu.
\newblock Ultralytics yolo11, 2024.
\newblock URL \url{https://github.com/ultralytics/ultralytics}.

\bibitem[Khan \& Fu(2023)Khan and Fu]{khan2023contrastive}
Zaid Khan and Yun Fu.
\newblock Contrastive alignment of vision to language through parameter-efficient transfer learning.
\newblock \emph{arXiv preprint arXiv:2303.11866}, 2023.

\bibitem[Koyun et~al.(2022)Koyun, Keser, Akkaya, and T{\"o}reyin]{koyun2022focus}
Onur~Can Koyun, Reyhan~Kevser Keser, Ibrahim~Batuhan Akkaya, and Beh{\c{c}}et~U{\u{g}}ur T{\"o}reyin.
\newblock Focus-and-detect: A small object detection framework for aerial images.
\newblock \emph{Signal Processing: Image Communication}, 104:\penalty0 116675, 2022.

\bibitem[Kuo et~al.(2022)Kuo, Cui, Gu, Piergiovanni, and Angelova]{kuo2022f}
Weicheng Kuo, Yin Cui, Xiuye Gu, AJ~Piergiovanni, and Anelia Angelova.
\newblock F-vlm: Open-vocabulary object detection upon frozen vision and language models.
\newblock \emph{arXiv preprint arXiv:2209.15639}, 2022.

\bibitem[Lam et~al.(2018)Lam, Kuzma, McGee, Dooley, Laielli, Klaric, Bulatov, and McCord]{lam2018xview}
Darius Lam, Richard Kuzma, Kevin McGee, Samuel Dooley, Michael Laielli, Matthew Klaric, Yaroslav Bulatov, and Brendan McCord.
\newblock xview: Objects in context in overhead imagery.
\newblock \emph{arXiv preprint arXiv:1802.07856}, 2018.

\bibitem[Li et~al.(2020)Li, Wan, Cheng, Meng, and Han]{li2020object}
Ke~Li, Gang Wan, Gong Cheng, Liqiu Meng, and Junwei Han.
\newblock Object detection in optical remote sensing images: A survey and a new benchmark.
\newblock \emph{ISPRS journal of photogrammetry and remote sensing}, 159:\penalty0 296--307, 2020.

\bibitem[Li et~al.(2024)Li, Guo, Yang, Liao, He, Zhou, and Yu]{li2024toward}
Yan Li, Weiwei Guo, Xue Yang, Ning Liao, Dunyun He, Jiaqi Zhou, and Wenxian Yu.
\newblock Toward open vocabulary aerial object detection with clip-activated student-teacher learning.
\newblock In \emph{European Conference on Computer Vision}, pp.\  431--448. Springer, 2024.

\bibitem[Li et~al.(2025)Li, Yang, Yang, Yan, Liu, Guo, and Chen]{li2025navblip}
Ye~Li, Li~Yang, Meifang Yang, Fei Yan, Tonghua Liu, Chensi Guo, and Rufeng Chen.
\newblock Navblip: a visual-language model for enhancing unmanned aerial vehicles navigation and object detection.
\newblock \emph{Frontiers in Neurorobotics}, 18:\penalty0 1513354, 2025.

\bibitem[Lin et~al.(2014)Lin, Maire, Belongie, Hays, Perona, Ramanan, Doll{\'a}r, and Zitnick]{lin2014microsoft}
Tsung-Yi Lin, Michael Maire, Serge Belongie, James Hays, Pietro Perona, Deva Ramanan, Piotr Doll{\'a}r, and C~Lawrence Zitnick.
\newblock Microsoft coco: Common objects in context.
\newblock In \emph{Computer vision--ECCV 2014: 13th European conference, zurich, Switzerland, September 6-12, 2014, proceedings, part v 13}, pp.\  740--755. Springer, 2014.

\bibitem[Lin et~al.(2017)Lin, Goyal, Girshick, He, and Doll{\'a}r]{lin2017focal}
Tsung-Yi Lin, Priya Goyal, Ross Girshick, Kaiming He, and Piotr Doll{\'a}r.
\newblock Focal loss for dense object detection.
\newblock In \emph{Proceedings of the IEEE international conference on computer vision}, pp.\  2980--2988, 2017.

\bibitem[Liu et~al.(2024{\natexlab{a}})Liu, Chen, Guan, Zhou, Zhu, Ye, Fu, and Zhou]{liu2024remoteclip}
Fan Liu, Delong Chen, Zhangqingyun Guan, Xiaocong Zhou, Jiale Zhu, Qiaolin Ye, Liyong Fu, and Jun Zhou.
\newblock Remoteclip: A vision language foundation model for remote sensing.
\newblock \emph{IEEE Transactions on Geoscience and Remote Sensing}, 2024{\natexlab{a}}.

\bibitem[Liu et~al.(2024{\natexlab{b}})Liu, Zeng, Ren, Li, Zhang, Yang, Jiang, Li, Yang, Su, et~al.]{liu2024grounding}
Shilong Liu, Zhaoyang Zeng, Tianhe Ren, Feng Li, Hao Zhang, Jie Yang, Qing Jiang, Chunyuan Li, Jianwei Yang, Hang Su, et~al.
\newblock Grounding dino: Marrying dino with grounded pre-training for open-set object detection.
\newblock In \emph{European Conference on Computer Vision}, pp.\  38--55. Springer, 2024{\natexlab{b}}.

\bibitem[Lu et~al.(2020)Lu, Li, Cui, Oswald, Pollefeys, and Qin]{lu2020geometry}
Xiaohu Lu, Zuoyue Li, Zhaopeng Cui, Martin~R Oswald, Marc Pollefeys, and Rongjun Qin.
\newblock Geometry-aware satellite-to-ground image synthesis for urban areas.
\newblock In \emph{Proceedings of the IEEE/CVF Conference on Computer Vision and Pattern Recognition}, pp.\  859--867, 2020.

\bibitem[Ma et~al.(2024)Ma, Zhang, Ding, Pun, and Ma]{ma2024decomposition}
Xianping Ma, Xiaokang Zhang, Xingchen Ding, Man-On Pun, and Siwei Ma.
\newblock Decomposition-based unsupervised domain adaptation for remote sensing image semantic segmentation.
\newblock \emph{IEEE Transactions on Geoscience and Remote Sensing}, 2024.

\bibitem[Miech et~al.(2020)Miech, Alayrac, Smaira, Laptev, Sivic, and Zisserman]{miech2020end}
Antoine Miech, Jean-Baptiste Alayrac, Lucas Smaira, Ivan Laptev, Josef Sivic, and Andrew Zisserman.
\newblock End-to-end learning of visual representations from uncurated instructional videos.
\newblock In \emph{Proceedings of the IEEE/CVF conference on computer vision and pattern recognition}, pp.\  9879--9889, 2020.

\bibitem[Minderer et~al.(2023)Minderer, Gritsenko, and Houlsby]{minderer2023scaling}
Matthias Minderer, Alexey Gritsenko, and Neil Houlsby.
\newblock Scaling open-vocabulary object detection.
\newblock \emph{Advances in Neural Information Processing Systems}, 36:\penalty0 72983--73007, 2023.

\bibitem[Mule et~al.(2025)Mule, Pannacci, Goudarzi, Pro, Papa, Maiano, and Amerini]{mule2025enhancing}
Emanuele Mule, Matteo Pannacci, Ali~Ghasemi Goudarzi, Francesco Pro, Lorenzo Papa, Luca Maiano, and Irene Amerini.
\newblock Enhancing ground-to-aerial image matching for visual misinformation detection using semantic segmentation.
\newblock In \emph{Proceedings of the Winter Conference on Applications of Computer Vision}, pp.\  795--803, 2025.

\bibitem[Mullappilly et~al.(2024)Mullappilly, Gehlot, Anwer, Khan, and Cholakkal]{mullappilly2024semi}
Sahal~Shaji Mullappilly, Abhishek~Singh Gehlot, Rao~Muhammad Anwer, Fahad~Shahbaz Khan, and Hisham Cholakkal.
\newblock Semi-supervised open-world object detection.
\newblock In \emph{Proceedings of the AAAI Conference on Artificial Intelligence}, volume~38, pp.\  4305--4314, 2024.

\bibitem[Qian et~al.(2022)Qian, Yang, Peng, Zhang, and Yan]{qian2022rsdet}
Wen Qian, Xue Yang, Silong Peng, Xiujuan Zhang, and Junchi Yan.
\newblock Rsdet++: Point-based modulated loss for more accurate rotated object detection.
\newblock \emph{IEEE Transactions on Circuits and Systems for Video Technology}, 32\penalty0 (11):\penalty0 7869--7879, 2022.

\bibitem[Radford et~al.(2021)Radford, Kim, Hallacy, Ramesh, Goh, Agarwal, Sastry, Askell, Mishkin, Clark, et~al.]{radford2021learning}
Alec Radford, Jong~Wook Kim, Chris Hallacy, Aditya Ramesh, Gabriel Goh, Sandhini Agarwal, Girish Sastry, Amanda Askell, Pamela Mishkin, Jack Clark, et~al.
\newblock Learning transferable visual models from natural language supervision.
\newblock In \emph{International conference on machine learning}, pp.\  8748--8763. PmLR, 2021.

\bibitem[Ranasinghe et~al.(2023)Ranasinghe, McKinzie, Ravi, Yang, Toshev, and Shlens]{ranasinghe2023perceptual}
Kanchana Ranasinghe, Brandon McKinzie, Sachin Ravi, Yinfei Yang, Alexander Toshev, and Jonathon Shlens.
\newblock Perceptual grouping in contrastive vision-language models.
\newblock In \emph{Proceedings of the IEEE/CVF International Conference on Computer Vision}, pp.\  5571--5584, 2023.

\bibitem[Redmon et~al.(2016)Redmon, Divvala, Girshick, and Farhadi]{redmon2016you}
Joseph Redmon, Santosh Divvala, Ross Girshick, and Ali Farhadi.
\newblock You only look once: Unified, real-time object detection.
\newblock In \emph{Proceedings of the IEEE conference on computer vision and pattern recognition}, pp.\  779--788, 2016.

\bibitem[Regmi \& Shah(2019)Regmi and Shah]{regmi2019bridging}
Krishna Regmi and Mubarak Shah.
\newblock Bridging the domain gap for ground-to-aerial image matching.
\newblock In \emph{Proceedings of the IEEE/CVF International Conference on Computer Vision}, pp.\  470--479, 2019.

\bibitem[Ren et~al.(2015)Ren, He, Girshick, and Sun]{ren2015faster}
Shaoqing Ren, Kaiming He, Ross Girshick, and Jian Sun.
\newblock Faster r-cnn: Towards real-time object detection with region proposal networks.
\newblock \emph{Advances in neural information processing systems}, 28, 2015.

\bibitem[Scheibenreif et~al.(2024)Scheibenreif, Mommert, and Borth]{scheibenreif2024parameter}
Linus Scheibenreif, Michael Mommert, and Damian Borth.
\newblock Parameter efficient self-supervised geospatial domain adaptation.
\newblock In \emph{Proceedings of the IEEE/CVF Conference on Computer Vision and Pattern Recognition}, pp.\  27841--27851, 2024.

\bibitem[Shan et~al.(2014)Shan, Wu, Curless, Furukawa, Hernandez, and Seitz]{shan2014accurate}
Qi~Shan, Changchang Wu, Brian Curless, Yasutaka Furukawa, Carlos Hernandez, and Steven~M Seitz.
\newblock Accurate geo-registration by ground-to-aerial image matching.
\newblock In \emph{2014 2nd International Conference on 3D Vision}, volume~1, pp.\  525--532. IEEE, 2014.

\bibitem[Shugaev et~al.(2024)Shugaev, Semenov, Ashley, Klaczynski, Cuntoor, Lee, and Jacobs]{shugaev2024arcgeo}
Maxim Shugaev, Ilya Semenov, Kyle Ashley, Michael Klaczynski, Naresh Cuntoor, Mun~Wai Lee, and Nathan Jacobs.
\newblock Arcgeo: Localizing limited field-of-view images using cross-view matching.
\newblock In \emph{Proceedings of the IEEE/CVF Winter Conference on Applications of Computer Vision}, pp.\  209--218, 2024.

\bibitem[Soto et~al.(2020)Soto, Costa, Feitosa, Happ, Ortega, Noa, Almeida, and Heipke]{soto2020domain}
PJ~Soto, GAOP Costa, RQ~Feitosa, PN~Happ, MX~Ortega, J~Noa, CA~Almeida, and Christian Heipke.
\newblock Domain adaptation with cyclegan for change detection in the amazon forest.
\newblock \emph{ISPRS Archives; 43, B3}, 43\penalty0 (B3):\penalty0 1635--1643, 2020.

\bibitem[Wang et~al.(2024{\natexlab{a}})Wang, Deng, Wang, Luo, Zhang, Zhang, and Jiang]{wang2024siamseg}
Bin Wang, Fei Deng, Shuang Wang, Wen Luo, Zhixuan Zhang, Gulan Zhang, and Peifan Jiang.
\newblock Siamseg: Self-training with contrastive learning for unsupervised domain adaptation semantic segmentation in remote sensing.
\newblock \emph{arXiv preprint arXiv:2410.13471}, 2024{\natexlab{a}}.

\bibitem[Wang et~al.(2024{\natexlab{b}})Wang, Wang, Li, Teng, and Li]{wang2024multi}
Kun Wang, Zi~Wang, Zhang Li, Xichao Teng, and Yang Li.
\newblock Multi-scale cross distillation for object detection in aerial images.
\newblock In \emph{European Conference on Computer Vision}, pp.\  452--471. Springer, 2024{\natexlab{b}}.

\bibitem[Wozniak et~al.(2024)Wozniak, Hansson, Thiel, and Jensfelt]{wozniak2024uada3d}
Maciej~K Wozniak, Mattias Hansson, Marko Thiel, and Patric Jensfelt.
\newblock Uada3d: Unsupervised adversarial domain adaptation for 3d object detection with sparse lidar and large domain gaps.
\newblock \emph{IEEE Robotics and Automation Letters}, 2024.

\bibitem[Yang et~al.(2019{\natexlab{a}})Yang, Fan, Chu, Blasch, and Ling]{yang2019clustered}
Fan Yang, Heng Fan, Peng Chu, Erik Blasch, and Haibin Ling.
\newblock Clustered object detection in aerial images.
\newblock In \emph{Proceedings of the IEEE/CVF international conference on computer vision}, pp.\  8311--8320, 2019{\natexlab{a}}.

\bibitem[Yang et~al.(2019{\natexlab{b}})Yang, Yang, Yan, Zhang, Zhang, Guo, Sun, and Fu]{yang2019scrdet}
Xue Yang, Jirui Yang, Junchi Yan, Yue Zhang, Tengfei Zhang, Zhi Guo, Xian Sun, and Kun Fu.
\newblock Scrdet: Towards more robust detection for small, cluttered and rotated objects.
\newblock In \emph{Proceedings of the IEEE/CVF international conference on computer vision}, pp.\  8232--8241, 2019{\natexlab{b}}.

\bibitem[Yang et~al.(2021)Yang, Yan, Feng, and He]{yang2021r3det}
Xue Yang, Junchi Yan, Ziming Feng, and Tao He.
\newblock R3det: Refined single-stage detector with feature refinement for rotating object.
\newblock In \emph{Proceedings of the AAAI conference on artificial intelligence}, volume~35, pp.\  3163--3171, 2021.

\bibitem[Yao et~al.(2024)Yao, Liu, Zhang, Ou, and Wu]{yao2024domain}
Liang Yao, Fan Liu, Chuanyi Zhang, Zhiquan Ou, and Ting Wu.
\newblock Domain-invariant progressive knowledge distillation for uav-based object detection.
\newblock \emph{IEEE Geoscience and Remote Sensing Letters}, 2024.

\bibitem[Ye et~al.(2024)Ye, He, Li, Lv, Yu, Yang, and He]{ye2024skydiffusion}
Junyan Ye, Jun He, Weijia Li, Zhutao Lv, Jinhua Yu, Haote Yang, and Conghui He.
\newblock Skydiffusion: Street-to-satellite image synthesis with diffusion models and bev paradigm.
\newblock \emph{arXiv preprint arXiv:2408.01812}, 2024.

\bibitem[Zareian et~al.(2021)Zareian, Rosa, Hu, and Chang]{zareian2021open}
Alireza Zareian, Kevin~Dela Rosa, Derek~Hao Hu, and Shih-Fu Chang.
\newblock Open-vocabulary object detection using captions.
\newblock In \emph{Proceedings of the IEEE/CVF conference on computer vision and pattern recognition}, pp.\  14393--14402, 2021.

\bibitem[Zeng et~al.(2024)Zeng, Gu, Qin, Jia, Deng, Xu, and Tian]{zeng2024unsupervised}
Junying Zeng, Yajin Gu, Chuanbo Qin, Xudong Jia, Senyao Deng, Jiahua Xu, and Huiming Tian.
\newblock Unsupervised domain adaptation for remote sensing semantic segmentation with the 2d discrete wavelet transform.
\newblock \emph{Scientific Reports}, 14\penalty0 (1):\penalty0 23552, 2024.

\bibitem[Zhang et~al.(2025)Zhang, Xiao, Yao, Zhang, and Gong]{zhang2025fusion}
Huiying Zhang, Pan Xiao, Feifan Yao, Qinghua Zhang, and Yifei Gong.
\newblock Fusion of multi-scale attention for aerial images small-target detection model based on pare-yolo.
\newblock \emph{Scientific Reports}, 15\penalty0 (1):\penalty0 4753, 2025.

\bibitem[Zhang et~al.(2019)Zhang, Yuan, Feng, and Lu]{zhang2019hierarchical}
Yuanlin Zhang, Yuan Yuan, Yachuang Feng, and Xiaoqiang Lu.
\newblock Hierarchical and robust convolutional neural network for very high-resolution remote sensing object detection.
\newblock \emph{IEEE Transactions on Geoscience and Remote Sensing}, 57\penalty0 (8):\penalty0 5535--5548, 2019.

\bibitem[Zhong et~al.(2022)Zhong, Yang, Zhang, Li, Codella, Li, Zhou, Dai, Yuan, Li, et~al.]{zhong2022regionclip}
Yiwu Zhong, Jianwei Yang, Pengchuan Zhang, Chunyuan Li, Noel Codella, Liunian~Harold Li, Luowei Zhou, Xiyang Dai, Lu~Yuan, Yin Li, et~al.
\newblock Regionclip: Region-based language-image pretraining.
\newblock In \emph{Proceedings of the IEEE/CVF conference on computer vision and pattern recognition}, pp.\  16793--16803, 2022.

\bibitem[Zhou et~al.(2022)Zhou, Girdhar, Joulin, Kr{\"a}henb{\"u}hl, and Misra]{zhou2022detecting}
Xingyi Zhou, Rohit Girdhar, Armand Joulin, Philipp Kr{\"a}henb{\"u}hl, and Ishan Misra.
\newblock Detecting twenty-thousand classes using image-level supervision.
\newblock In \emph{European conference on computer vision}, pp.\  350--368. Springer, 2022.

\bibitem[Zhu et~al.(2020)Zhu, Su, Lu, Li, Wang, and Dai]{zhu2020deformable}
Xizhou Zhu, Weijie Su, Lewei Lu, Bin Li, Xiaogang Wang, and Jifeng Dai.
\newblock Deformable detr: Deformable transformers for end-to-end object detection.
\newblock \emph{arXiv preprint arXiv:2010.04159}, 2020.

\end{thebibliography}
